\pdfoutput=1
\documentclass[preprint,5p,twocolumn]{elsarticle}

\usepackage{amssymb}
\usepackage{amsmath}
\usepackage{graphicx}
\usepackage{caption}
\usepackage{subcaption}
\usepackage{color}

\newcommand{\Var}[1]{\operatorname{Var}\left[#1\right]}
\newcommand{\etal}{{et al.}}

\journal{Signal Processing : Image Communications}

\begin{document}

\begin{frontmatter}



\title{A Unified Framework for Multi-Sensor HDR Video Reconstruction}


\author[link]{Joel Kronander\corref{cor1}}
\cortext[cor1]{Corresponding Author}
\ead{joel.kronander@liu.se}

\author[link]{Stefan Gustavson}

\author[sph]{Gerhard Bonnet}

\author[link]{Anders Ynnerman}

\author[link]{Jonas Unger}

\address[link]{C-Research, Department of Media and Information Technology, Link\"oping University}
\address[sph]{AG Spheron VR}

\begin{abstract}
One of the most successful approaches to modern high quality HDR-video capture is to use camera setups with multiple sensors imaging the scene through a common optical system.
However, such systems pose several challenges for HDR reconstruction algorithms. Previous reconstruction techniques have considered debayering, denoising, resampling (alignment) and exposure fusion as separate problems. In contrast, in this paper we present a unifying approach, performing HDR assembly directly from raw sensor data. Our framework includes a camera noise model adapted to HDR video and an algorithm for spatially adaptive HDR reconstruction based on fitting of local polynomial approximations to observed sensor data. The method is easy to implement  and allows reconstruction to an arbitrary resolution and output mapping. We present an implementation in CUDA and show real-time performance for an experimental 4 Mpixel multi-sensor HDR video system. We further show that our algorithm has clear advantages over existing methods, both in terms of flexibility and reconstruction quality.
\end{abstract}


\end{frontmatter}


\section{Introduction}

\emph{High dynamic range} (HDR) video is an emerging field of technology, by many considered to be one of the key components of future imaging, that will enable a wide range of new applications in image processing, computer vision, computer graphics and cinematography. A few prototype HDR-video camera systems have been presented recently, showing that the computational power and bandwidth is now high enough to handle high resolution HDR-video processing and storage. Current CMOS and CCD imaging sensors are still unable to accurately capture the full dynamic range in general scenes. There are sensors with logarithmic response to incident light that exhibit significantly higher dynamic range compared to standard CMOS and CCD sensors~\cite{Technologies:uq}. These are, however, in most applications still not accurate enough due to problems with image noise and that the entire dynamic range is usually quantized to 10-12 bit output. To date, the most successful approach for high quality HDR-video capture has instead been to use camera setups with multiple sensors imaging the scene through a common optical system~\cite{Aggarwal2004,Mcguire2007,Tocci2011,Kronander:2013fk}. The idea is similar to exposure bracketing~\cite{Debevec1998}, but instead of varying the exposure time, the optical setup is designed to make the sensors capture different exposures of the scene by the use of different neutral density (ND) filters. By capturing all exposures simultaneously, temporal artifacts such as ghosting and incorrect motion blur inherent to techniques where the exposure time is varied, e.g. \cite{Kang2003, Unger2007}, are avoided. Although multi-sensor systems currently provide the (arguably) best alternative for high quality HDR video capture, there has been a lack of a formalized framework for efficient and accurate image reconstruction for general sensor configurations. 

In order to reconstruct HDR images from multi-sensor systems it is necessary to perform: \emph{demosaicing} of color filter array (CFA) sampled data, \emph{resampling} to correct for geometric misalignments between the sensors, \emph{HDR assembly} to fuse the input low dynamic range (LDR) images into the output HDR image, and \emph{denoising} to reduce image and color noise.

\begin{figure}[!h]
\begin{center}
\includegraphics[width=\linewidth]{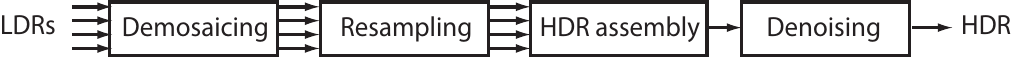}
\end{center}
\caption{\label{fig:hdrpipeline}The vast majority of previous HDR reconstruction methods have considered \emph{demosaicing}, \emph{resampling}, \emph{HDR assembly} and \emph{denoising} as separate problems.}
 \end{figure}

Most existing HDR reconstruction techniques, e.g.~\cite{Debevec1998,Ajdin2008,Granados2010,Tocci2011}, have considered this as separate steps in a pipeline fashion and performed demosaicing and realignment either before or after HDR assembly, see Figure~\ref{fig:hdrpipeline}. This pipeline approach introduces several problems. Demosaicing before HDR assembly as in~\cite{Debevec1998}, causes problems with bad or missing data around saturated pixels. This is especially problematic as color channels usually saturate at different levels. Demosaicing after HDR assembly as in~\cite{Ajdin2008,Tocci2011} causes problems with blur and ghosting unless the sensors are perfectly aligned. For multi-sensor systems using high resolution sensors, it is problematic, costly, and sometimes even impossible to match the sensors so that the CFA patterns align correctly. \citet{Tocci2011} report misalignment errors around the size of a pixel despite considerable alignment efforts. This means that overlapping pixels from different sensors might not have the same spectral response. Also, treating demosaicing, resampling and HDR assembly in separate steps makes it difficult to analyze and incorporate sensor noise in the HDR reconstruction in a coherent way. 

The main contribution of this article is a filtering framework that addresses all of these steps simultaneously in a formalized way. 
Our reconstruction method takes into account the entire set of non-saturated LDR samples available around the output HDR pixel, and performs demosaicing, resampling, HDR assembly and denoising in a single operation. The framework is based on spatially adaptive cross-sensor filtering, and uses a noise-aware local polynomial approximation (LPA) approach adapted to include heterogeneous noise in the reconstruction. Using our framework, we present two flexible and robust algorithms for high quality HDR reconstructions using both isotropic filtering support and adaptive anisotropic filtering support with correlations between the color channels taken into account. A key feature of our framework is its computational speed, making it applicable to real-time HDR reconstruction for video applications. Using isotropic filtering, our CUDA implementation of the reconstruction algorithm runs at a sustained rate of 25 fps for input data from a 4 Mpixel system with four sensors.  As a part of our framework, we have also developed a new sensor noise model.

To demonstrate the usefulness of our approach, we evaluate the algorithms and noise model using both simulated data for a range of possible multi-sensor setups for HDR-imaging as well as on data from an experimental multi-sensor HDR-video camera prototype.

\section{Related Work}
To the authors' knowledge, no previous approach has considered HDR reconstruction with joint realignment, noise reduction and CFA interpolation. While our method is not designed to provide optimal results with respect to denoising, demosaicing or super-resolution alone, we take a unified approach based on a model of sensor noise and provide a flexible framework which is straightforward to implement and use. Our method is also well suited to the needs of a versatile high dynamic range video system, by allowing fast parallel processing and an arbitrary resolution for the reconstruction. In comparison to our previous work~\cite{Kronander:2013fk}, where we only consider isotropic filtering support, we use adaptive anisotropic filtering support to adapt the reconstruction to the local image structure. This also enables us to incorporate correlations between the color channels in the reconstruction, similar to modern demosaicing methods~\cite{Gunturk2005}.

The algorithms presented in this paper are related to a large body of previous work, ranging from HDR capture and reconstruction, e.g.~\cite{reinhard2010high,MYSZKOWSKI:2008fk}, to theory and algorithms for accurate image reconstruction and image fusion, e.g.~\cite{Astola:2006fk}.
In this section, we give an overview of the previous work most closely related to the methods proposed in this paper. 

\subsection{HDR Video Capture}
\citet{Nayar2000} proposed an HDR imaging system suitable for video capture, based on placing a neutral density (ND) filter array over the image sensor introducing a tradeoff between spatial resolution and dynamic range. While early methods only considered grayscale images the approach was  later generalized to include spectral sampling using a spatially varying exposure and color filter array~\cite{Narasimhan2005,Yasuma2010}. 
Solutions to the sampling and resolution enhancement of image dimensions such as spatial, spectral, temporal and intensity information have also been solved previously using multiple sensors for the same optical viewpoint. A simple example is the tri-color CCD system, where a beam splitter separates the color spectrum into three components (R, G, B) which are then measured individually on three different sensors. Aggarwal and Ahuja~\cite{Aggarwal2004} present a system for HDR video capture using a mirror pyramid aligned with the optical axis to split incoming light and consider different exposure times or ND-filters for the sensors, or they consider mirrors with an offset from the optical axis to effectively distribute light unevenly over the sensors. Mcguire et al.~\cite{Mcguire2007} presented a framework for optimally selecting components for optical splitting trees given specific target goals, such as dynamic range, spectral sensitivity and cost budgets. Tocci \etal~\cite{Tocci2011} recently presented a light efficient multi-sensor HDR video system utilizing up to 99.96\% of the incident light, which has spawned a renewed interest in these systems.
Hybrid approaches could also use a multi-sensor setup where one of the sensors includes a spatially varying ND-filter array. Our reconstruction method is general and applies to all of these various setups.

\subsection{HDR fusion}
The most common method for HDR reconstruction from a set of differently exposed Low Dynamic Range (LDR) images is to compute a per-pixel weighted average of the LDR measurements. The weights, often based on heuristics, are chosen to suppress image noise and remove saturated values from processing~\cite{Mann:1995zr,Debevec1998,Akyuz:2007ys} .
Weight functions can also be based on more sophisticated camera noise models. Mitsunaga and Nayar~\cite{Mitsunaga1999} derived a weight function that maximizes SNR assuming signal-independent additive noise. Kirk and Andersen~\cite{Kirk2006} derived a weight function inversely proportional to the temporal variance of the digital LDR values. \citet{Granados2010} later extended this approach to include both spatial and temporal camera noise.
While most previous methods consider only a single pixel at a time from each LDR exposure, \citet{Tocci2011} presented an algorithm that incorporates a neighborhood of LDR samples in the reconstruction. This method helps smooth the transition regions between sensors and performs well for sensors with large exposure differences, over 3 \textit{f}-stops apart. 

\subsection{Non-parametric Image Processing}
The last two decades have seen an increased popularity of image processing operations using locally adaptive filter weights, for applications in e.g. interpolation, denoising and upsampling. Examples include normalized convolution~\cite{Knutsson1993}, the bilateral filter~\cite{Tomasi1998}, and moving least squares~\cite{Lancaster:1981nx}. Recently, deep connections have been shown~\cite{Takeda2007,Milanfar:2011kx} between these methods and traditional non-parametric statistics~\cite{Loader:1999uq}. In this paper, we fit Local Polynomial Approximations (LPA)~\cite{Astola:2006fk} to irregularly distributed samples around output pixels using a localized maximum likelihood estimation~\cite{Tibshirani:1987fk} to incorporate the heterogeneous noise of the samples.

\subsection{Image Fusion}
When multiple images of the same scene are captured within sub-pixel displacements of each other, super-resolution reconstruction methods can be applied.  A class of computationally simple super-resolution techniques are those based on interpolation in a common high resolution grid~\cite{SuperRes:2010uq}. These methods typically first use some registration to put observed images in a common high resolution reference frame. The high-resolution image is then estimated in a least squares sense given a local neighborhood of observed pixels~\cite{Farsiu:2004bh,Hardie:2007vn}.
As noted in previous work, super-resolution and HDR imaging complement each other. Super-resolution increases the resolution in the spatial domain, and HDR increases the radiometric range and resolution. Several methods have also been proposed to incorporate both enhancements in the same processing framework~\cite{Gunturk:2006ve,Zimmer:2011ly}. 

\subsection{Demosaicing}  
Most modern camera systems acquire color images using a sensor covered by a Color Filter Array (CFA), and they estimate the final color vector for each pixel from neighboring color samples. This process is known as demosaicing. The most frequently used CFA of today is the RGB Bayer pattern~\cite{BayerPatent76}, sampling the G channel on a quincunx grid and the R and B images on a rectangular grid at half the native resolution of the sensor. Several demosaicing methods have been developed through the years. A comprehensive overview can be found in Gunturk \etal~\cite{Gunturk2005}. Joint denoising and demosaicing have been considered in previous work~\cite{K.-Hirakawa:2006uq,Hirakawa:2008kx}. Using a global maximum a posteriori optimization with priors that enforce correlation between color channels, \citet{farsiu2006multiframe} proposed a framework for joint super-resolution and demosaicing. A similar approach was taken in~\cite{Anton-Kachatou:2008fk} for HDR imaging with joint demosaicing for sensors with non-destructive readout (i.e. perfect alignment between consecutive readouts/frames). A global MAP optimization can also be used for joint HDR fusion, realignment and demosaicing. However, it is computationally demanding and does not straightforwardly allow for parallel processing, making it less suitable for the HDR video applications considered in this paper.

\begin{figure}[t]
\centering
\includegraphics[width=\linewidth]{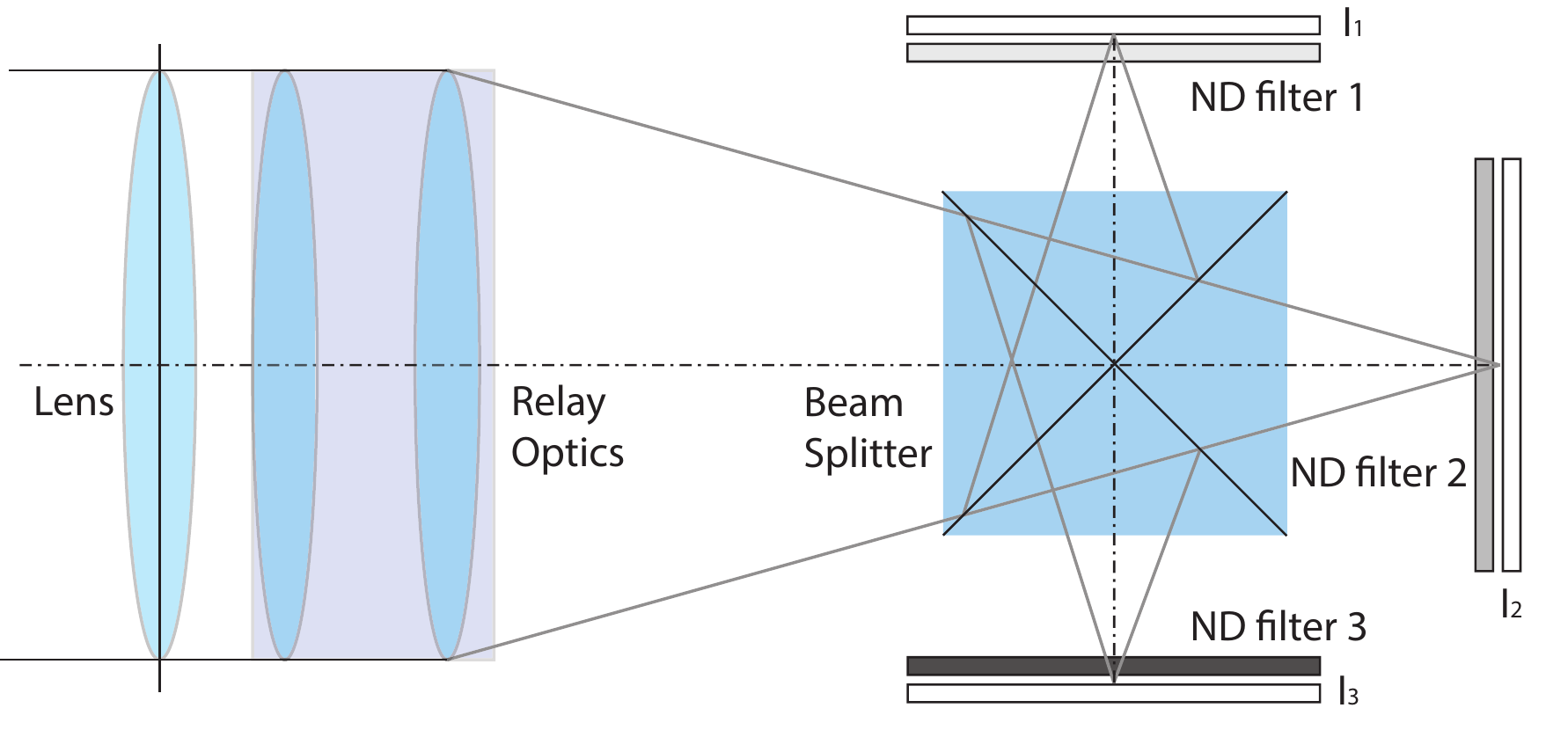}
\caption{Sketch of a general multi-sensor setup. 
Each sensor, $I_s$ receives a fraction, $n_s$ of the incident light depending on the beamsplitter and ND-filter configuration.
\label{fig:opticalsetup}
}
\end{figure}

\begin{figure}[t]
\centering
\includegraphics[width=\linewidth]{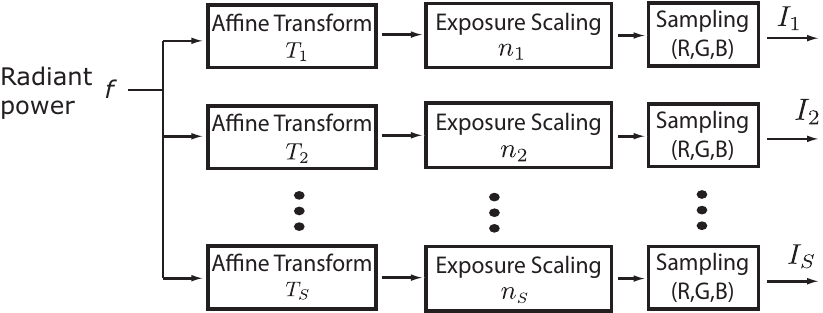}
\caption{The image formation model used to describe the HDR capture.
\label{fig:imageformationmodel}
}
\end{figure}

\section{Problem formulation}
The goal of the algorithm presented in this paper is to generate a video stream of output HDR images based on input data from multi-sensor imaging systems which for each output frame capture a set of $N_s$ differently exposed raw CFA images $I_s$, where $s=1...N_{s}$. The algorithm is designed for multi-sensor systems where the sensors image the scene through a common front lens, and where the light path is split onto the sensors by a beam splitter arrangement. An example multi-sensor setup is displayed in Figure~\ref{fig:opticalsetup}. We assume that misalignments between the sensors can be described as 2-D transformations, $T_{s}$. The difference in exposure between the sensors, created by e.g. neutral density (ND) filters placed in the ray path, can be described as an exposure scaling coefficient $n_s$ for each sensor. Each pixel in the input images provides a measurement of the incident radiant power at the pixel location, scaled by an unknown factor due to the quantum efficiency of the sensor, \cite{Hasinoff:2010zr}. In this work we assume that the sensors have the same quantum efficiency or that it can be measured separately for each sensor so that it can be incorporated into the exposure scaling coefficient $n_s$. 

This yields the image formation model described in Figure~\ref{fig:imageformationmodel}. Each sensor image, $I_{s}$, samples the incident radiant power, $f$, at a set of discrete pixel locations. Using a linear index $i$ for pixels in each sensor image, we define the measured digital sample value at a pixel $i$ in image $I_{s}$ as $y_{s,i}$. The samples, $y_{s,i}$, contain measurement noise that is dependent on the incident radiant power, $f_{s,i}$, as well as the sensor characteristics. 

During capture, we assume that the sensors are synchronized or that the scene is static, so that the motion blur characteristics are the same for all sensors. The parameters that are allowed to vary between the sensors are: the exposure time $t_s$,  the sensor gain setting $g_s,$ and the exposure scaling coefficient, $n_s$.

\begin{figure}
\begin{center}
\centering
\includegraphics[width=\linewidth]{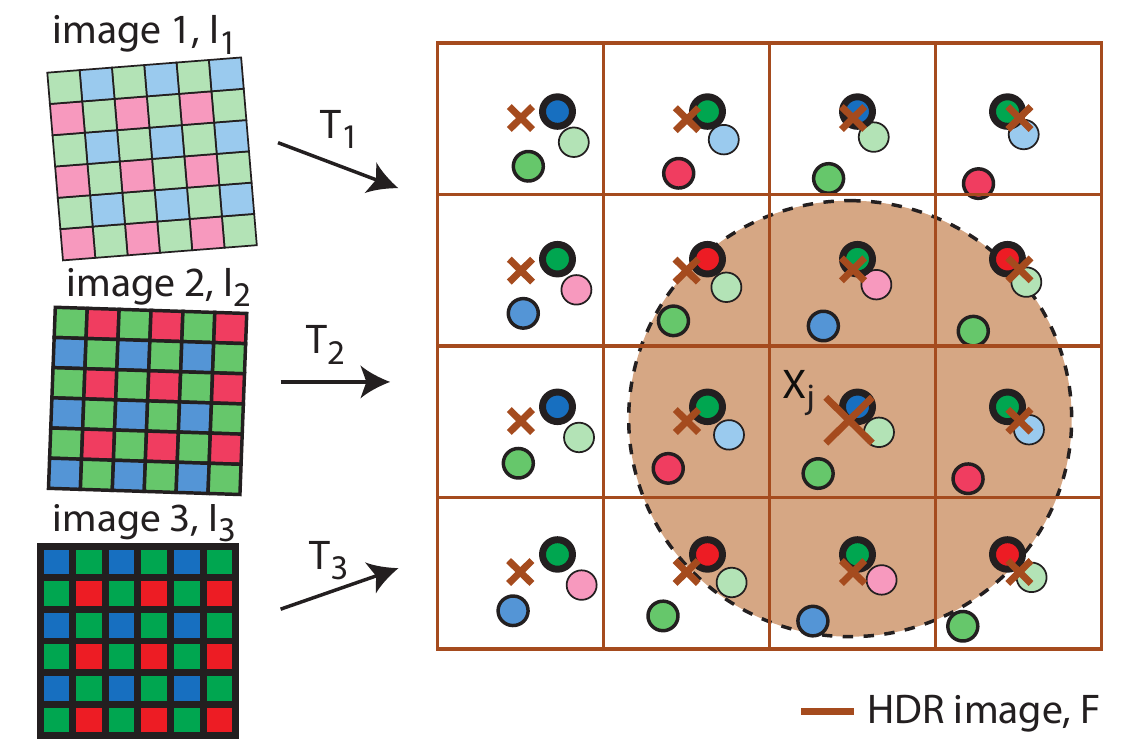}
\caption{HDR reconstruction is performed to a virtual reference system with arbitrary resolution and mapping. The sensor samples are mapped to the reconstruction grid using the transformations $T_{s}$. The pixel at position $X_j$  in the HDR image is estimated using a local polynomial approximation of nearby samples (within the shaded circle). Saturated and defective pixels are discarded,  and samples near the black level, local outliers and samples farther from the reconstruction point contribute less to the radiance estimate.
\label{fig:registeredframes}
}
\end{center}
\vspace{-15pt}
\end{figure}

\section{Calibration}
\label{sec:algorithm}
Before estimating the HDR frames, we perform radiometric and geometric calibration of the sensors.

\subsection{Radiometric Calibration}
The (noisy) digital pixel values $y_{s,i}$ in the input images, $I_{s}$, are first converted to radiant power estimates $\hat{f}_{s,i}$. This is carried out by subtracting a bias frame and  taking into account the exposure settings (exposure time, exposure scaling, global gain) and per-pixel gain variations.  The variance of the radiant power, $\sigma^2_{\hat{f}_{s,i}}$, can then be estimated using a calibrated radiometric camera noise model. Similarly to~\cite{Granados2010}, we assume the noise to follow a Gaussian distribution. This assumption is described in detail in Section~\ref{subsec:sensornoisemodel}, where we give an in-depth overview of our radiometric noise model adapted for HDR video. It should be noted that our framework can use any radiometric noise model that assumes an approximately Gaussian noise distribution.

\subsection{Geometric Alignment}
\label{sec:geomtransforms}
The HDR reconstruction is performed in a virtual reference space, corresponding to a virtual sensor placed somewhere in the focal plane. The virtual sensor dimensions are chosen to reflect the desired output frame, unconstrained by the resolution of the input frames. Using a geometric calibration procedure for each sensor an affine transform, $T_s$, is established which maps sensor pixel coordinates, $x_{s,i}$, to the coordinates of the reference output coordinate system, $X_{s,i} = T_s(x_{s,i})$. In general, the transformed input pixel coordinates, $X_{s,i}$, are not integer-valued in the output image coordinates, and for general affine transformations the sample locations $X_{s,i}$ will become irregularly positioned. An example for three sensor images is shown in Figure~\ref{fig:registeredframes}.

\section{HDR Reconstruction}
\label{sec:Greconstruction}
To reconstruct the HDR output frame the relative radiant power is estimated for each pixel in the virtual sensor separately in a non-parametric fashion using the transformed samples $\hat{f}_{s,i}(X_{s,i})$. For each output pixel $j$ with integer valued, regularly spaced coordinates $X_{j}$, a local polynomial model is fitted to observed samples $\hat{f}_{s,i}(X_{s,i})$ in a local neighborhood, see Figure~\ref{fig:registeredframes}. We first discuss how a single color channel, $c = \{R || G || B\}$, can be reconstructed independently of the other channels using LPA with isotropic filtering supports. In section~\ref{sec:AdaptiveLPArec} it is then shown how the first LPA estimate can be improved by using adaptive anisotropic filtering support. Using adaptive filtering support also enables us to utilize cross-correlation between color channels to improve the estimate. This is discussed in section~\ref{sec:colorchannelcorr}.

\subsection{Local Polynomial Model}
To estimate the radiant power, $f(x)$, at an output pixel, we use a generic local polynomial expansion of the radiant power around the output pixel location $X_j = [x_1, x_2]^T$. Assuming that the radiant power $f(x)$ is a smooth function in a local neighborhood around the output location $X_j$ a \emph{M{\small-th}} order Taylor series expansion is used to predict the radiant power at a point $X_i$ close to $X_j$ as:
\begin{align}
\widetilde{f}(X_i) &= C_0 + C_1(X_i-X_j) \nonumber \\ 
& + C_2 tril\{(X_i-X_j)(X_i-X_j)^T\} + \quad ... 
\end{align}
where \emph{tril} lexicographically vectorizes the lower triangular part of a symmetric matrix and where 
\begin{align}
C_0 &= f(X_j) \\
C_1 &= \nabla f(X_j) = \bigg{[}\frac{\partial f(X_j)}{\partial x_{1}}, \frac{\partial f(X_j)}{\partial x_{2}}\bigg{]}\\
C_2 &= \frac{1}{2}\bigg{[} \frac{\partial^2 f(X_j)}{\partial x_{1}^2}, 2\frac{\partial^2 f(X_j)}{\partial x_{1} \partial x_2}, \frac{\partial^2 f(X_j)}{\partial x_{2}^2} \bigg{]}
\end{align} 
Given the fitted polynomial coefficients, $C_{1:M}$, we can thus predict the radiant power at the output location $X_j$ by $C_0 = f(X_j)$, and the first order gradients by $C_1$. In this work we only consider local expansions of order $M \leq 2$.

\subsection{Maximum Localized Likelihood Fitting}
To estimate the coefficients, $C$, of the local polynomial model, we maximize a localized likelihood function~\cite{Tibshirani:1987fk} defined using a smoothing window centered around $X_{j}$:
\begin{align}
\mathcal{W}_H(X) = \frac{1}{det(H)}\mathcal{W}(H^{-1}X)
\end{align}
where $H$ is a $2 \times 2$ smoothing matrix that determines the shape and size of the window. In section~\ref{sec:windowselection}, we discuss how the shape and size of the window function and smoothing can be selected.
To lessen the notational burden, the observed samples in the neighborhood, $\{ \hat{f}_{s,i}(X_{s,i}) : X_{s,i} \in \operatorname{supp} \big{(}\mathcal{W}_H(X) \big{)} \}$, are denoted below by $f_k$ with a linear index $k = 1... K$. Using the assumption of normally distributed radiant power estimates $f_k$, see Section~\ref{subsec:sensornoisemodel}, the log of the localized likelihood for the polynomial expansion centered at $X_{j}$ is given by
\begin{align}
& L(X_{j},C) = \sum_{k}{\log({N(f_k|\widetilde{f}(X_k),\hat{\sigma}_{f_{k}}^2)\mathcal{W}_H(X_k)})}  \nonumber \\
&= \sum_{k}{}\bigg{[}f_{k}-C_0-C_1(X_i-X_j) \nonumber \\
& - C_2^Ttril\{(X_i-X_j)(X_i-X_j)^T\} - ...\bigg{]}^2\frac{\mathcal{W}_H(X_k)}{\hat{\sigma}_{f_{k}}} +R \nonumber
\end{align}
where $R$ represents terms independent of $C$.
The polynomial coefficients, $\tilde{C}$, maximizing the localized likelihood function is found by the weighted least squares estimate
\begin{align}
\tilde{C} &= \underset{C \in \mathcal{R}^M}{\operatorname{argmax}}(L(X_{j},C)) \nonumber \\
&= (\Phi^TW\Phi)^{-1}\Phi^TW\bar{f} \label{eq:wlsqe}
\end{align}
where
\begin{align}
\bar{f} &= [f_1, f_2, ... f_K]^T \nonumber \\
W &= diag[\frac{\mathcal{W}_H(X_1)}{\hat{\sigma}_{f_1}}, \frac{\mathcal{W}_H(X_2)}{\hat{\sigma}_{f_2}}, ..., \frac{\mathcal{W}_H(X_K)}{\hat{\sigma}_{f_K}}] \nonumber \\
\Phi &= \begin{bmatrix} 
1 & (X_1-X_j) & tril^T\{(X_1-X_j)(X_1-X_j)^T\} & ... \\ 
1 & (X_2-X_j) & tril^T\{(X_2-X_j)(X_2-X_j)^T\} & ... \\
\vdots & \vdots & \vdots & \vdots \\
1 & (X_K-X_j) & tril^T\{(X_K-X_j)(X_K-X_j)^T\} & ...
 \end{bmatrix} \nonumber
\end{align}
Note that similarly to the kernel regression framework~\cite{Takeda2007}, using equivalent kernels, the solution to the weighted least squares problem can also be formulated as a locally adaptive linear filtering.

\subsection{Parameters}
\label{subsec:parameters}
The expected mean square error of the reconstructed image depends on a trade-off between bias and variance of the estimate. This trade-off is determined by: the order of the polynomial basis $M$, the window function $\mathcal{W}$, and the smoothing matrix $H$. 


\begin{figure}
\begin{center}
\begin{subfigure}[b]{0.5\linewidth}
         \centering
	\includegraphics[width=\textwidth]{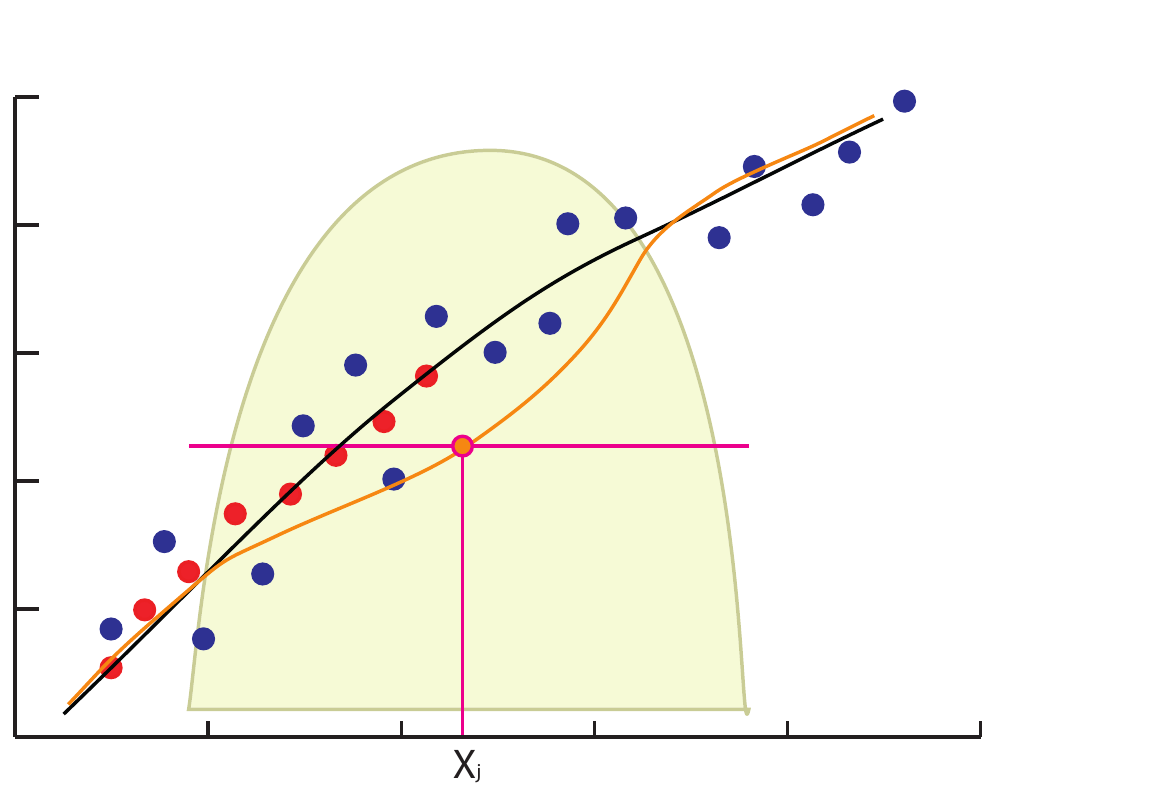}
	\caption{Using a constant fit.}
	\label{subfig:localfit}
\end{subfigure}%
\begin{subfigure}[b]{0.5\linewidth}
	\centering
	\includegraphics[width=\textwidth]{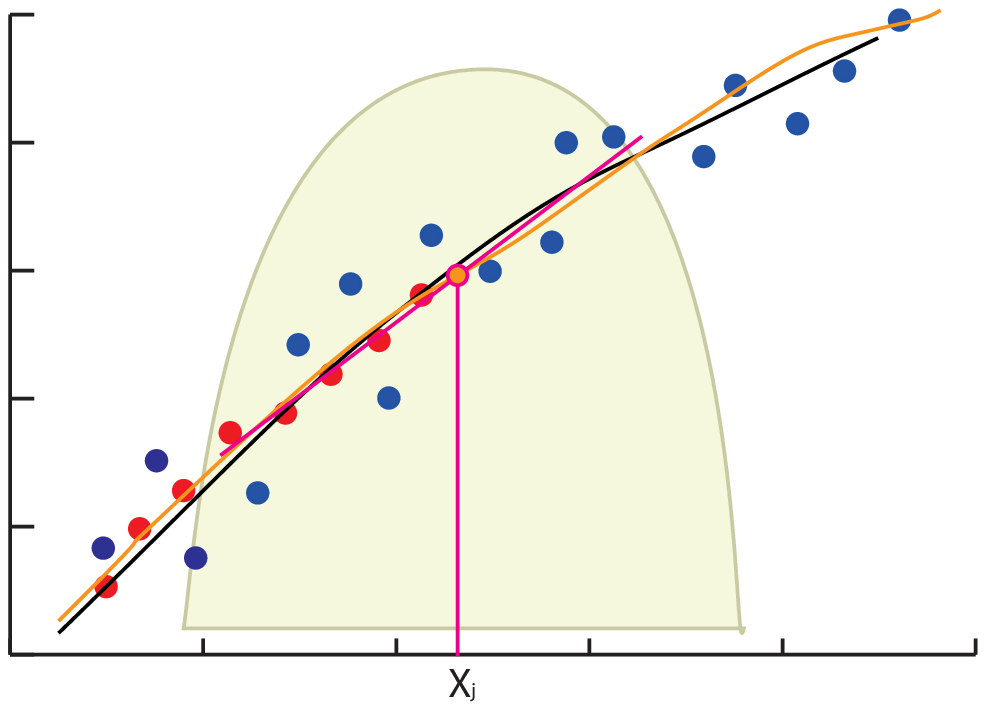}
	\caption{Using a linear fit.}
	\label{subfig:linearfit}
\end{subfigure}%
\caption{1D example showing the effect of the polynomial order, $M$, in areas affected by sensor saturation. The red circles represent a sensor saturating at lower values than the sensor represented by the blue circles. \subref{subfig:localfit}) Using a constant fit, $M=0$, introduces significant bias as more unsaturated measurements are available to the left of the reconstruction point. \subref{subfig:linearfit}) Using a linear fit, $M=1$, the bias is reduced drastically.}
\label{fig:polyfit}
\end{center}
\vspace{-10pt}
\end{figure}

Using a piecewise constant polynomial, $M = 0$, the estimator corresponds to an ordinary locally weighted average of neighboring sensor observations
\begin{equation}
\hat{f}(X_{j})   =  \frac{\sum_{k} {\frac{\mathcal{W}_H(X_k)}{\hat{\sigma}_{f_k}}f_k}} { \sum_{k} {\frac{\mathcal{W}_H(X_k)}{\hat{\sigma}_{f_k}}}}
\end{equation}
However, locally weighted averages can exhibit severe bias at image boundaries and in regions around sensor saturation, due to the asymmetry of the number of available sensor measurements in these locations.
By instead fitting a linear polynomial (a plane, $M=1$), the bias can be reduced significantly, see Figure~\ref{fig:polyfit}. Introducing higher order polynomials is possible, but may lead to increased variance in the estimates. In this work, we only consider $M \leq 2$.

\subsubsection{Window and scale selection}
\label{sec:windowselection}
In this paper we only show results using a Gaussian window function
\begin{align}
\mathcal{W}_H(X_k) = \frac{1}{2\pi det(H)}exp\bigg{\{} -(X_k - X_j)^TH^{-1}(X_k - X_j)\bigg{\}}
\end{align}
There are also other possible choices of symmetric smooth window functions, e.g. Epanechnikov or Tricube windows~\cite{Astola:2006fk}. In this work we consider the Gaussian window function for its simplicity and widespread use.

The $2 \times 2$ smoothing matrix, $H$, affects the shape of the Gaussian. It is important that the support of the smoothing window is extended to incorporate enough samples into the estimate in Equation~\eqref{eq:wlsqe} so that $(\Phi^TW\Phi)$ is invertible (full rank). 
The simplest choice of the smoothing matrix is $H = hI$, where $h$ is a global scale parameter and I is the identity matrix. This corresponds to an isotropic filter support. The choice of $h$ is dependent on the sensor characteristics (noise) and the scene. We therefore treat $h$ as a user parameter. 
To reconstruct sharp images in low noise conditions, we generally choose the smallest possible $h$ such that $(\Phi^TW\Phi)$ is invertible. However, if the captured images exhibit severe noise, a larger scale parameter may be beneficial. For Bayer CFA sensors, we consider different scale parameters for the green and red/blue color channels, $h_G = \frac{h_{R,B}}{\sqrt{2}}$, as there are more green samples per unit area. For the remainder of this paper we will refer to the reconstruction method using a global smoothing matrix, $H = hI$, as  \emph{Local Polynomial Approximation (LPA)}.

\subsection{Adaptive LPA Reconstruction}
\label{sec:AdaptiveLPArec}
The truncated local polynomial expansion assumes that the radiant power, $f(x)$, is a smooth function (up to order $M$) in a neighborhood of the reconstruction point $X_j$. In many image regions these assumptions are not valid. For example, if an output pixel is located near a sharp edge, it  cannot be represented accurately with a finite polynomial expansion. It is therefore desirable to adapt the window support so that it only includes observations on the same side of the edge.

To adapt the support of the local polynomial model to the data, we use a two step approach: First, we use LPA with $M \geq 1$ and isotropic window support to compute an initial estimation of the local gradients, $\bigg{[}\frac{\partial f(X_i)}{\partial x_{1}}, \frac{\partial f(X_i)}{\partial x_{2}}\bigg{]}$. Using the estimated gradients, we then locally adapt the smoothing matrix, $H_j$, to the local structure of the signal at each output pixel $X_j$. 

We define an oriented, anisotropic smoothing matrix as $H_j^s = hC_j$ where $C_j$ represents the local gradient covariance of the signal estimated from the data. To estimate $C_j$, we use a parametric approach similar to~\citet{Takeda2007}. Specifically, we consider a decomposition of the covariance matrix as:
\begin{align}
C_j &= \gamma_j U_{\theta_j} \Lambda_j U_{\theta_j} \\
U_{\theta_j} &= \begin{bmatrix} 
cos(\theta_j) & sin(\theta_j) \\
-sin(\theta_j) & cos(\theta_j)
\end{bmatrix}\\
\Delta_j &= \begin{bmatrix} 
\sigma_j & 0 \\
0 & \frac{1}{\sigma_j}
\end{bmatrix}
\end{align}
This corresponds to describing the covariance matrix as a function of three parameters: $\sigma_j$ describing an elongation along the principal directions, $\theta_i$ describing a rotation angle and $\gamma_i$ describing an overall scaling. The elongation, rotation and scaling parameters are estimated from a truncated singular value decomposition of weighted gradients in a local neighborhood around the output location $X_j$:
\begin{align}
G = \begin{bmatrix}
\vdots & \vdots \\
f^{x_1}(X_i) & f^{x_2}(X_i) \\
\vdots & \vdots
\end{bmatrix}
= U_jS_jV_j \quad \forall X_i \in supp\{w_i\}
\end{align}
where $f^{x_1}(X_i)$ and $f^{x_2}(X_i)$ are the estimated gradients along $x_1$ and $x_2$ at the nearby point $X_i$, and $w_i$ represents a local window function. The singular values $s_1$ and $s_2$ along the diagonal of $S_j$ represent the energy along the dominant directions in the neighborhood defined by the local window function. Thus, the orientation angle is set so that the window is elongated orthogonal to the dominant gradient direction, i.e. from the second column of $V_j$, $v = [v_1, v_2]^T$ as
\begin{align}
\theta_j = arctan\bigg{(}\frac{v_1}{v_2}\bigg{)}
\end{align}
The elongation parameter, $\sigma_j$ is then set according to:
\begin{align}
\sigma_j = \frac{s_1 + \lambda_1}{s_2 + \lambda_1}, \lambda_1 \geq 0
\end{align}  
where $\lambda_1$ is a regularization parameter.
The scaling parameter, $\gamma_j$, is finally set according to 
\begin{align}
\gamma_i = \bigg{(} \frac{s_1s_2 + \lambda_2}{M} \bigg{)}^\alpha
\end{align}
where $\lambda_2$ is another regularization parameter, $\alpha$ is the structure sensitivity parameter, and $M$ is the number of samples in the local neighborhood of the gradient analysis window $w$.
Intuitively the shape of the window support should be: circular and relatively large in flat areas, elongated along edges, and small in textured areas to preserve detail. For all results presented in this paper we fix the regularization parameters to $\lambda_1 = 1$ and $\lambda_2 = 0.001$. The structure sensitivity parameter, $\alpha$, is treated as a user specified parameter, representing a tradeoff between denoising and detail preservation. For a more detailed analysis of the choice of parameters, we refer the reader to~\citet{Takeda2007} and \citet{Milanfar:2011kx}. Note that as we are primarily interested in irregularly sampled data we choose to adapt the window functions instead of the support of each observation (kernel) as in the work of Takeda \etal

\subsection{Color Channel Correlation}
\label{sec:colorchannelcorr}
Modern demosaicing methods consider the correlation between color channels to improve CFA interpolation. Commonly used heuristics are that edges should match between color channels and that interpolation should be performed along and not across edges. For Bayer pattern CFA arrays, in which the G channel is sampled more densely than the R and B channels, a common method is to first reconstruct the G channel, then use the gradients of the G channel when reconstructing the R and B channel. 

To take into account the correlation between color channels in our reconstruction, we adapt this idea to the irregularly sampled input data from multi-sensor camera systems. First we estimate the gradients for the G channel using LPA with $M  \geq  1$ and isotropic window supports. We then use the G channel gradients to locally adapt $H_j$ as described in the previous section. Using the adapted $H_{j}$, we then compute the radiant power estimates for all color channels, (R, G, B). This effectively forces the interpolation to be performed along and not across edges and preserves their location between the color channels. We refer to this reconstruction method as \emph{Color Adaptive Local Polynomial Approximation (CALPA)}.

\section{Sensor noise model} 
\label{subsec:sensornoisemodel} 
In this section we describe a radiometric camera noise model adapted to HDR video.
We assume a linear digital response and model the noise using a radiometric camera model 
taking into account the exposure times, gain settings and ND-filter coefficients for the different sensors.  
Our noise model is inspired by previous methods considering radiometric camera calibration~\cite{JAP:8102423,Foi:2008ys} and optimal HDR capture methods~\cite{Granados2010,Hasinoff:2010zr}. 

Each observed digital pixel value, $y_{s,i}$, is obtained using an exposure time, $t_s$, and an exposure scaling, $n_s$ that is assumed constant over the image $I_s$.
The pixel response is a measurement of the radiant power reaching the image sensor, which we for convenience express as the number of photo-induced electrons collected per unit time, $f_{s,i}$.
The accumulated number of photoelectrons, $e_{s,i}$, collected at the pixel during the exposure time, $t_s$, follows a Poisson distribution with expected value
\begin{align}
E[e_{s,i}] = t_{s}(a_{s,i}n_{s}f_{s,i})
\end{align}
and variance $\Var{e_{s,i}} = E[e_{s,i}]$ where $a_{s,i}$ is a per pixel factor due to non-uniform photo-response. The recorded digital value, $y_{s,i}$, is also affected by the sensor gain and signal independent readout noise, 
\begin{align}
	y_{s,i} = g_s(e_{s,i}) + r_{s,i}(g)
\end{align}
where $g_s$ is the analog amplifier/sensor gain (proportional to the native ISO settings on modern cameras) and $r_{s,i}(g)$ is the readout noise which generally depends on the gain setting of the sensor. As we focus on video applications with frame rates of $25$ fps or more, the dark current noise is neglected. For most modern camera sensors dark current noise has negligible effect for exposures less than a second~\cite{Hasinoff:2010zr}. In contrast to previous work~\cite{Hasinoff:2010zr,Kronander:2013fk} we do not assume a simple parametric model for the readout noise dependence on the gain, as we have found that such models do not generally provide a satisfactory fit to measured readout noise at different gain settings for modern camera sensors. To handle sensors with different gain settings, we instead calibrate the readout noise, $r_{s,i}(g)$ for each separate gain/ISO setting considered.

Before reconstruction of the HDR output frame, each digital pixel value, $y_{s,i}$, is independently transformed to an estimate of the number of photoelectrons reaching the pixel per unit time, $\hat{f}_{s,i}$. To compensate for the readout noise bias (blacklevel), $E[d_{s,i}(g,t)]$, we subtract a \emph{bias frame}, $b_{s,i}$, from each observation. The bias frame is computed as the average of a large set of black images captured with the same camera settings as the observations but with the lens covered, so that no photons reach the sensor. The radiant power $f_{s,i}$ can then be estimated as
\begin{align}
\hat{f}_{s,i} = \frac{y_{s,i} - b_{s,i}} {g_st_sa_{s,i}n_s}\label{eq:Efsi}
\end{align}

\subsection{Variance estimate}
We assume $\hat{f}_{s,i}$ to follow a normal distribution with mean $f_{s,i}$ and standard deviation $\sigma_{\hat{f}}$. For low light levels photon shot noise is generally dominated by signal independent readout noise approximately following a normal distribution. In contrast for brighter regions the Poisson distributed shot noise is well approximated by a normal distribution. Assuming no saturated or clipped pixel values, the variance of $\hat{f}_{s,i}$ is given by
\begin{align}
\Var{\hat{f}_{s,i}} = \sigma_{\hat{f}}^2 =  \frac{g_s^2(\Var{e_{s,i}}) + \Var{r_{s,i}(g,t)}}{g_s^2t_s^2a_{s,i}^2n_s^2}
\end{align}
Due to the photon shot noise, to estimate, $\Var{e_{s,i}}$, we should ideally know the true value of $f_{s,i}$, however in practice we are instead forced to use the (noisy) estimate, $\hat{f}_{s,i}$, obtained from equation~\eqref{eq:Efsi}.
Thus we form an approximate variance estimate, $\hat{\sigma}_{\hat{f_{s,i}}}^2$, as
\begin{align}
\hat{\sigma}_{\hat{f_{s.i}}}^2 &= \frac{g_s^2t_sa_{s,i}n_{s}\hat{f} + Var[r_{s,i}(g,t)]}{g_s^2t_s^2a_{s,i}^2n_s^2}  \label{eq:varianceestiamte}
\end{align}

The analysis above assumes non-saturated pixels. In practice, we exclude saturated pixels by comparing each digital pixel value, $y_{s,i}$, to a \textit{saturation frame}. The model does not include the effect of pixel cross-talk, and the variances, $\sigma_{\hat{f_{s,i}}}^2$, are assumed to be independent of each other.
More complicated models of sensor noise also include the effects of saturation on the digital value variance, $\Var{y_{s,i}}$, see e.g.~\cite{Foi:2009fk}. However such models assume $y_{s,i}$ to follow a clipped normal distribution which requires an impractical non-linear estimation when computing the coefficients for the local polynomial approximations discussed in section~\ref{sec:Greconstruction}.

\subsection{Parameter calibration}
\label{subsection:parametercalibration}
The exposure scaling, $n_{s}$, and per-pixel non-uniformity, $a_{s,i}$, can in practice be estimated using a flat field image computed as the average over a large sequence of images, running the sensors at different exposure times ensure well exposed pixels for all sensors. Choosing one sensor as reference, the exposure scaling, $n_{s}$, can be estimated as the spatial mean of the ratio, computed pixel-wise between the radiant power estimates, Eq.~\ref{eq:Efsi}, of the sensors. The non-uniformity, $a_{s,i}$, can then be computed as the per-pixel deviation from the exposure scaling, $n_{s}$.

The variance of the readout noise, $\Var{r_{s,i}(g,t)}$ can be estimated, similarly to the bias frame, from a set of black images, captured with the same camera settings, ($g$, $t$), as the observations but with the lens covered, so that no photons reach the sensor.

The sensor gain, $g_{s}$, can be calibrated using the relation,
\begin{align}
\frac{\Var{y_{s,i}}-\Var{b_{s,i}}}{E[y_{s,i}]-E[b_{s,i}]}  = \frac{g_{s}^2\Var{e_{s,i}}}{g_{s}E[e_{s,i}]} = g_{s}
\end{align}
where the second equality follows from $e_{s,i}$ being Poisson distributed shot noise with $E[e_{s,i}] = \Var{e_{s,i}}$. 
$E[y_{s,i}]$ and $E[b_{s,i}]$ can be estimated by averaged flat fields and the bias frame respectively, and $\Var{b_{s,i}}$ as described above.


\label{sec:noisevalidation}


\begin{figure*}[!t]
\begin{center}
\begin{subfigure}[b]{0.55\linewidth}
\centering
\includegraphics[width=1\linewidth]{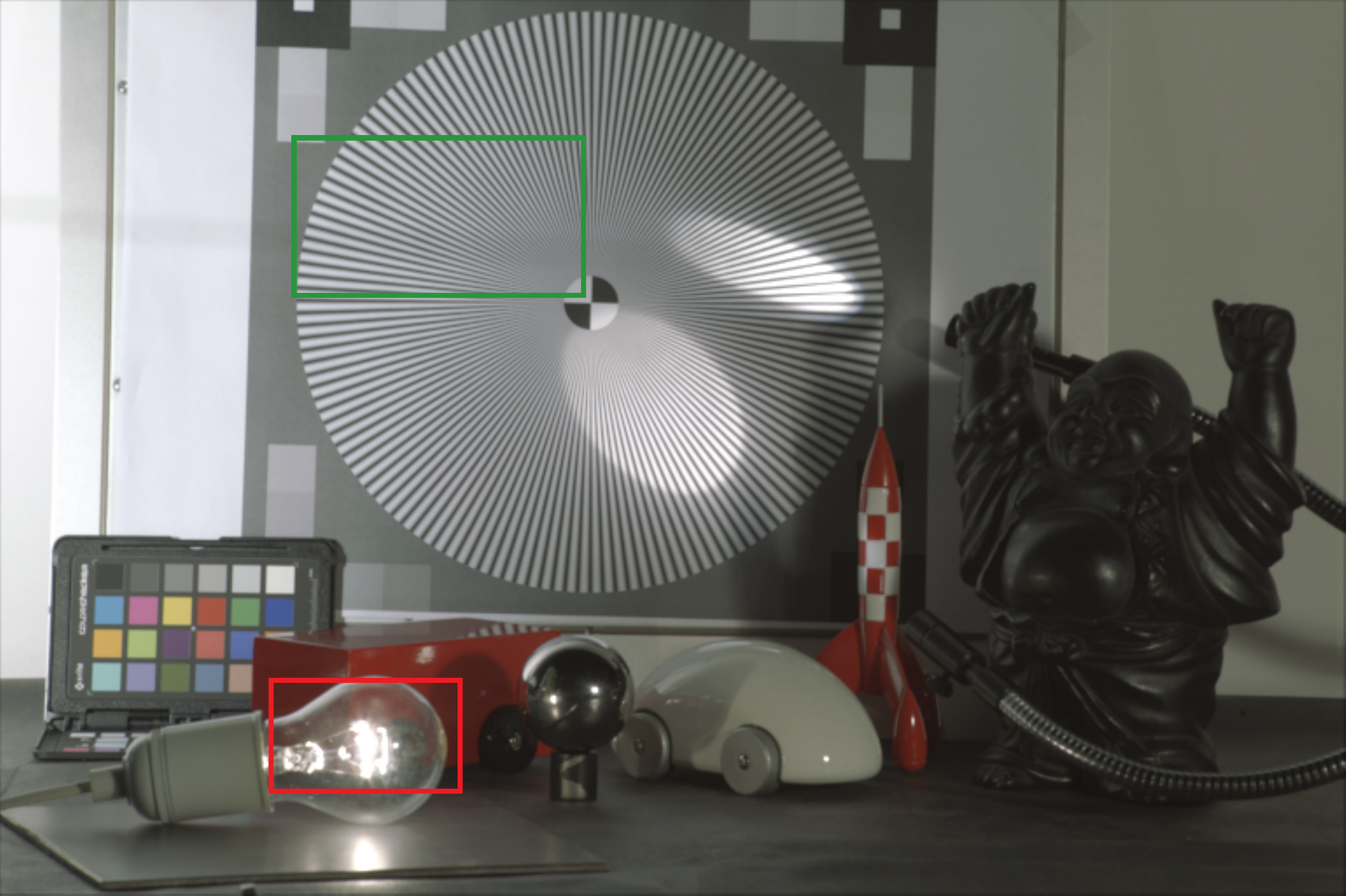}
\caption{HR Reference}
\end{subfigure}

\begin{subfigure}[b]{0.285\linewidth}
\includegraphics[width=1\linewidth]{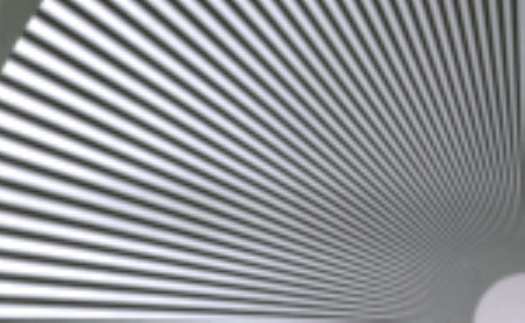}
\subcaption{Reference}
\includegraphics[width=1\linewidth]{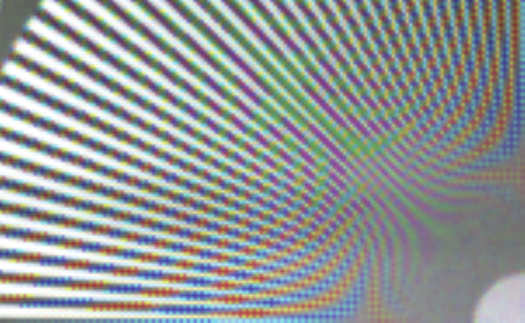}
\subcaption{Debayer First, Bilinear}
\end{subfigure}
\begin{subfigure}[b]{0.285\linewidth}
\includegraphics[width=1\linewidth]{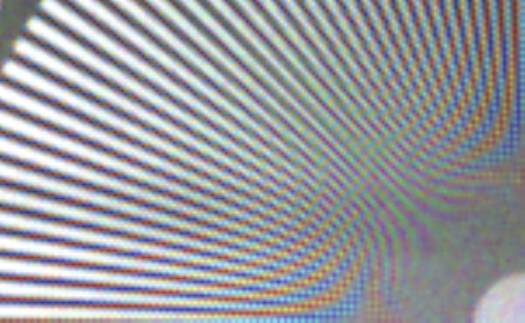}
\subcaption{LPA}
\includegraphics[width=1\linewidth]{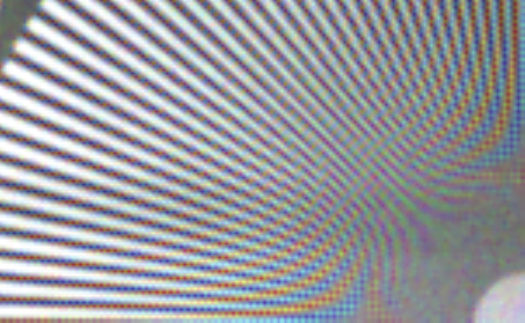}
\subcaption{Debayer Last, Bilinear}
\end{subfigure}
\begin{subfigure}[b]{0.285\linewidth}
\includegraphics[width=1\linewidth]{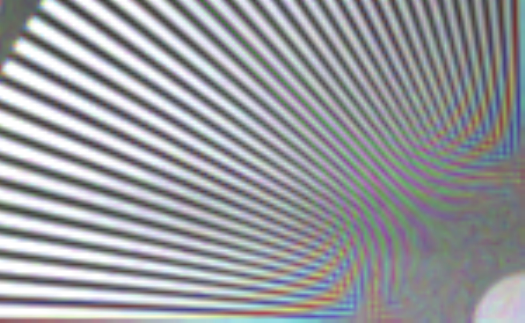}
\subcaption{Color Adaptive LPA}
\includegraphics[width=1\linewidth]{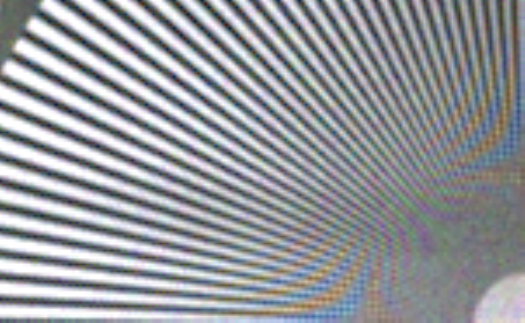}\subcaption{Debayer Last, using \cite{Malvar:2004fk}}
\end{subfigure}

\begin{subfigure}[b]{0.285\linewidth}
\includegraphics[width=1\linewidth]{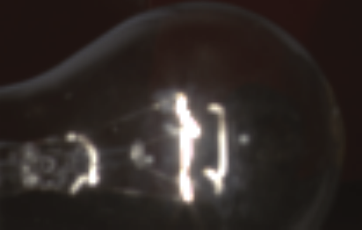}
\subcaption{Reference}
\includegraphics[width=1\linewidth]{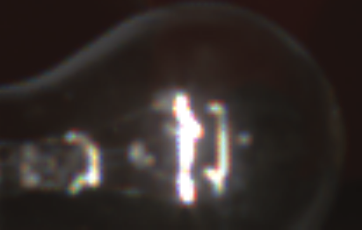}
\subcaption{Debayer First, Bilinear}
\end{subfigure}
\begin{subfigure}[b]{0.285\linewidth}
\includegraphics[width=1\linewidth]{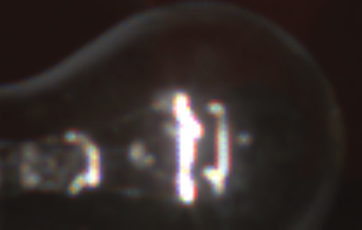}
\subcaption{LPA}
\includegraphics[width=1\linewidth]{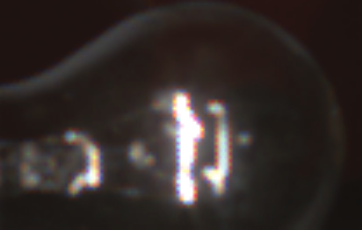}
\subcaption{Debayer Last, Bilinear}
\end{subfigure}
\begin{subfigure}[b]{0.285\linewidth}
\includegraphics[width=1\linewidth]{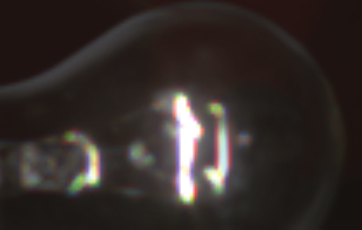}
\subcaption{Color Adaptive LPA}
\includegraphics[width=1\linewidth]{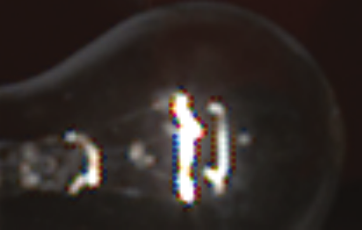}
\subcaption{Debayer Last, using \cite{Malvar:2004fk}}
\end{subfigure}
\caption{Reconstructed results from 3 perfectly aligned virtual Canon 5D sensors captured 3 f-stops apart.}
\label{fig:PerfectAlignmentComparison}
\end{center}
\vspace{-10pt}
\end{figure*}

\section{Algorithm evaluation}
\label{sec:simulateddata}
To evaluate our algorithm, we compare its performance against several reconstruction methods. As most previously proposed algorithms do not generalize to setups with arbitrary misalignments between the sensors, we include evaluations on setups where the sensors are assumed to be perfectly aligned as well as setups where the sensors only have a translational offset from each other. 

We compare our algorithms with the following methods:
\begin{itemize}
\item{ \emph{Tocci \etal} : The recent real-time multi-sensor reconstruction algorithm presented by \citet{Tocci2011}.}
\item{ \emph{Debayer-first} : A method performing demosaicing before HDR assembly, using heuristic linear radiometric weights.}
\item{ \emph{Debayer-last:} A method performing demosaicing after HDR assembly, using radiometric weights inversely proportional to the variance estimate $\hat{\sigma}_{\hat{f}}^2$ given by our sensor noise model~\ref{subsec:sensornoisemodel}.}
\item{\emph{Granados \etal} : The iterative method proposed by Granados~\etal~\cite{Granados2010} for high-quality HDR assembly of perfectly aligned data.}
\end{itemize}
For the above methods we consider demosaicing using both bilinear interpolation and the gradient adaptive method presented by \citet{Malvar:2004fk}.

We base the majority of our comparisons on HDR image data generated from a camera simulation framework. In the simulator, the scene is represented by a virtually noise free high resolution HDR image, meticulously generated from a set of exposures captured 1 \emph{f}-stop apart covering the dynamic range of the real scene. Each exposure is computed as an average of $100$ frames, in this case captured with a Canon 5D SLR camera. The camera simulation framework allows us to generate images of arbitrary bit depth, sensor misalignments, exposure settings and noise characteristics according to the model described in Section~\ref{subsec:sensornoisemodel}. For our experiments, we simulate different multi-sensor setups  with different sensor misalignments, $T_{s}$, and different noise characteristics, given the simulation parameters $g_s, n_s, t_s, Var[r_{s,i}]$.
We focus our evaluation on simulated sensors with three to four f-stops difference between the sensor exposures as this is the most common setup for multi-sensor HDR video systems in practice~\cite{Tocci2011}.

\subsection{Perfect alignment}
Using perfectly aligned sensors allows us to compare our more general algorithms to a range of previous methods only applicable to this type of sensor data. We note that these sensor setups are not very realistic for multi-sensor video applications. However, we show that our methods yield image quality that is comparable and in some cases even better than state-of-the-art high quality reconstruction methods even for these cases.
In Figure~\ref{fig:PerfectAlignmentComparison} the local polynomial approximation (LPA) and color adaptive local polynomial approximation (CALPA) methods are compared to the \emph{Debayer-first} method and the \emph{Debayer-last} method. The virtual sensors were perfectly aligned, captured with exposure scalings $n_1 = 1, n_2 = 2^{-4}, n_3= 2^{-8}$ and set to simulate the noise properties of a Canon 5D sensor, with parameters: $g_s = 0.23$, and $\Var{r_{s,i}} = 6.5$ as reported by~\cite{Granados2010}. The Debayer-first method performs poorly for high frequency details, introducing significant color artifacts. The Debayer-last method performs slightly better for bilinear CFA interpolation. Using the gradient adaptive method of \citet{Malvar:2004fk} effectively suppresses color moir\'e artifacts in low contrast regions, however in high contrast regions, such as close to the light bulb filament, significant color artifacts are introduced. Compared to the Debayer-first method, the LPA reconstruction reduces color artifacts by performing resampling and CFA interpolation jointly with reconstruction. LPA reconstruction also gracefully handles transitional regions around sensor saturation points, as a smoothing spatial filtering and a cross-sensor blending are both inherent in the reconstruction. This has the benefit of implicitly performing denoising. Using anisotropic window supports, correlated between the color channels, CALPA produces even smoother results in high contrast regions, e.g. close to the filament. CALPA also produces less color artifacts than LPA in regions with high spatial frequency. 

\begin{figure}[!t]
\begin{center}
\begin{subfigure}[b]{0.49\linewidth}
\centering
\includegraphics[width=1\linewidth]{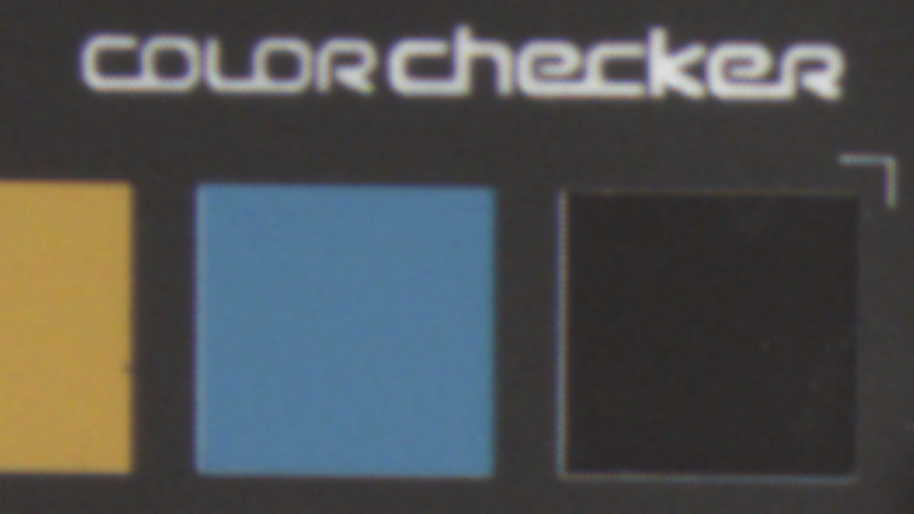}
\caption{LPA, $M=1$, $h_{R/B} = 0.7$}
\end{subfigure}
\begin{subfigure}[b]{0.49\linewidth}
\centering
\includegraphics[width=1\linewidth]{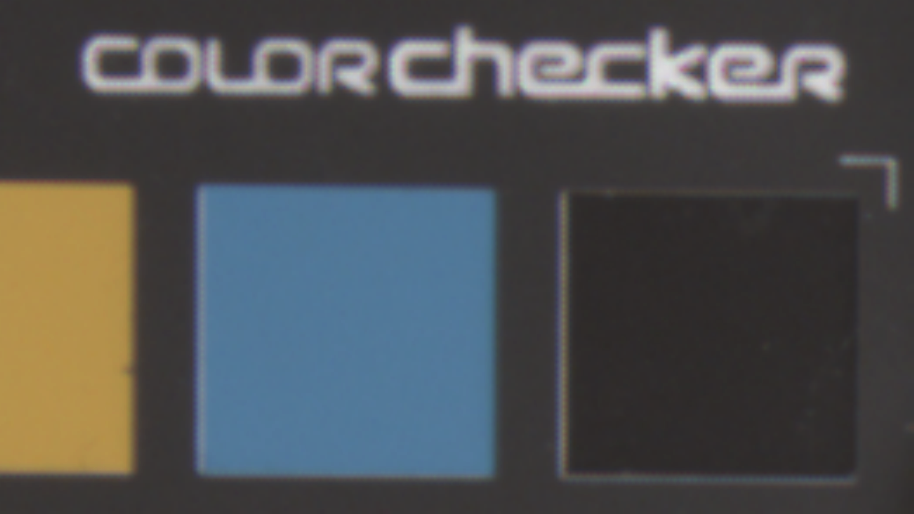}
\caption{Granados \etal~\cite{Granados2010}}
\end{subfigure}
\caption{Reconstructed results from four perfectly aligned exposures 4 f-stops apart captured with a Canon 40D camera. (a) LPA Reconstruction and (b) the iterative method of Granados \etal~\cite{Granados2010}.}
\label{fig:comparisonGranados}
\end{center}
\vspace{-25pt}
\end{figure}

To compare our method to the algorithm proposed by~\citet{Granados2010} we captured four exposures 4 f-stops apart with a Canon 40D camera, with calibrated gain $1.24$ (ISO 400), and readout noise variance $64.2$ (14 bit sensor). For a fair comparison to our method we perform demosaicing (bilinear CFA interpolation) after HDR-assembly for Granados method, in contrast to the publicly available code~\cite{hdrwebpage} that performs a crude nearest neighbor demosaicing before HDR assembly. Careful inspection of the reconstructions, see e.g. Figure~\ref{fig:comparisonGranados}, shows that the method proposed by~\citet{Granados2010} performs slightly better, although the difference is not striking. Note that in contrast to the method of~\citet{Granados2010} our method handles arbitrary misalignments between the sensors.

\begin{figure*}[!t]
\begin{center}
\begin{subfigure}[b]{0.22\linewidth}
 \includegraphics[width=\linewidth]{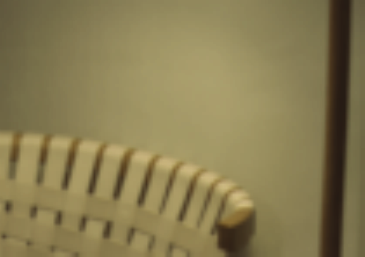}
 \includegraphics[width=\linewidth]{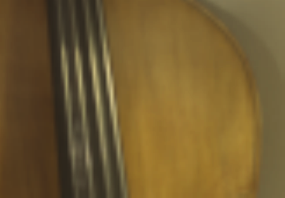}
 \subcaption[b]{Reference}
\label{subfig:comparison3LPAM0}
\end{subfigure}
\begin{subfigure}[b]{0.22\linewidth}
 \includegraphics[width=\linewidth]{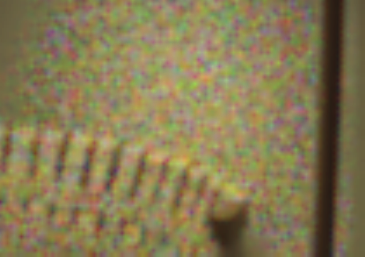}
 \includegraphics[width=\linewidth]{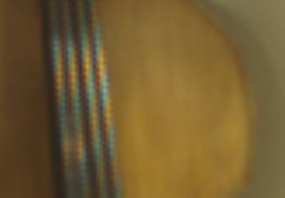}
 \subcaption[b]{\citet{Tocci2011}}
\label{subfig:comparison3LPAM0}
\end{subfigure}
\begin{subfigure}[b]{0.22\linewidth}
 \includegraphics[width=\linewidth]{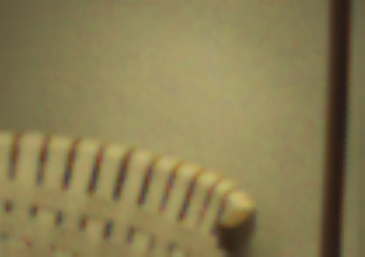}
 \includegraphics[width=\linewidth]{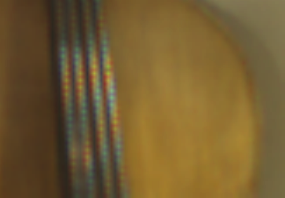}
 \subcaption[b]{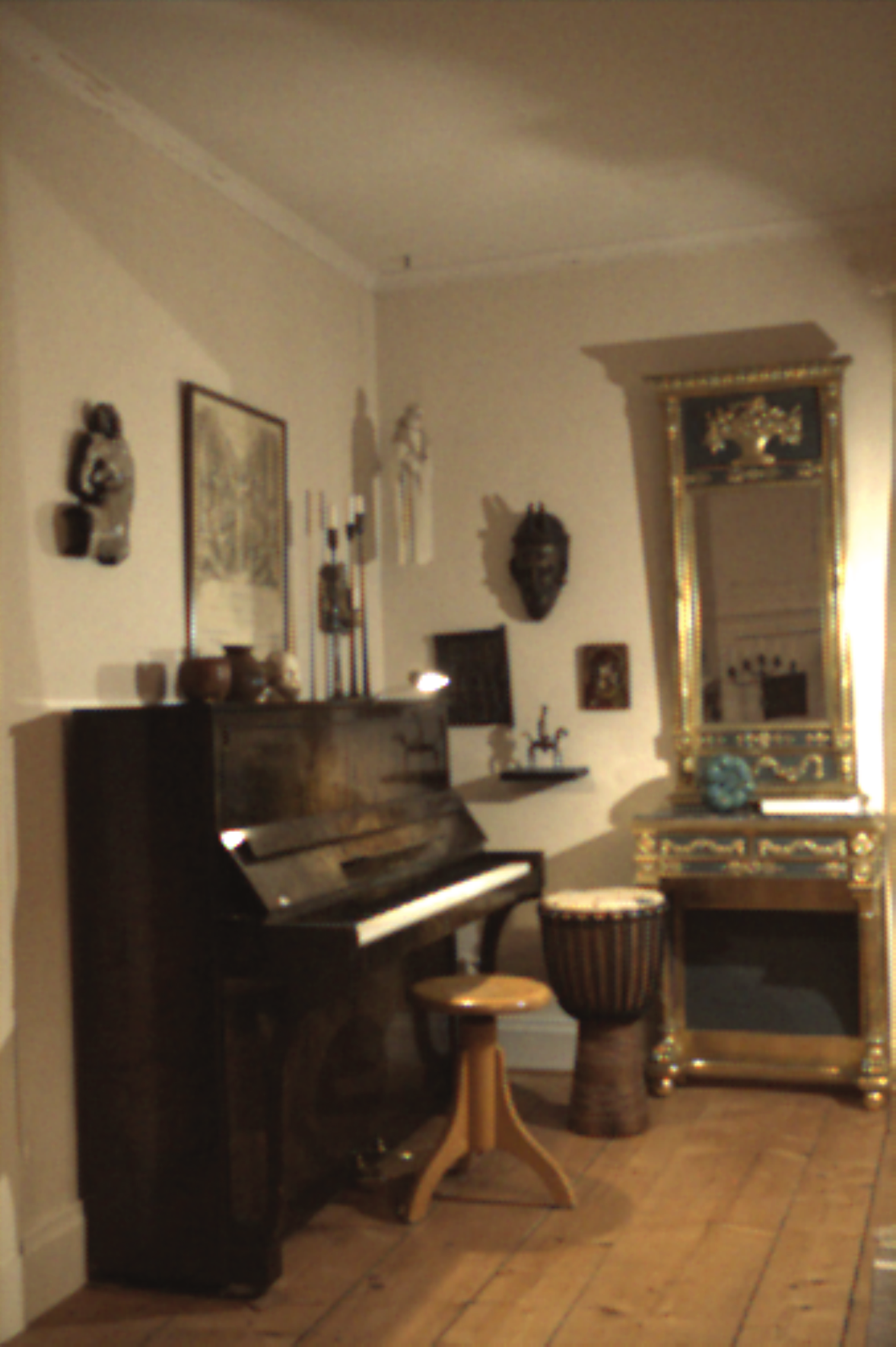}
\label{subfig:comparison3LPAM0}
\end{subfigure}
\begin{subfigure}[b]{0.22\linewidth}
 \includegraphics[width=\linewidth]{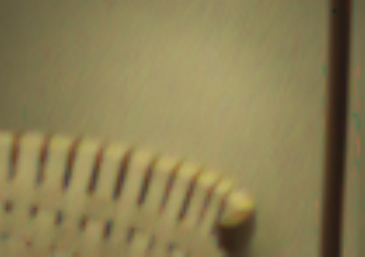}
 \includegraphics[width=\linewidth]{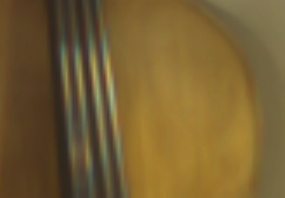}
 \subcaption[b]{CALPA}
\label{subfig:comparison3LPAM0}
\end{subfigure}

\caption{Magnifications from reconstructions of a scene virtually exposed with 3 Kodak Kai-04050 sensors 5 f-stops apart. a) Tonemapped HDR reference images. b) Reconstructions by the method proposed by \citet{Tocci2011}, c)LPA $M = 1$, $h = 0.7$, d)CALPA, $M = 1$, $h = 0.7$, $\alpha = 0.005$. In areas of sensor transitions, the method by Tocci \etal exhibits severe noise as only one sensor is used at a time, as can be seen in the magnification on the top row. In contrast LPA handles sensor transitions gracefully due to the incorporation of a sensor noise model and a statistical maximal likelihood fit to all sensor data. CALPA show ever better results due to the inherent denoising using adaptive filtering supports. Using CALPA also help in reducing unwanted color artifacts.}
\label{fig:translationalmissalignment}
\end{center}
\end{figure*}

\subsection{Sensor Missalignment}
 In Figure~\ref{fig:translationalmissalignment}, LPA and CALPA are compared to the real-time multi-sensor reconstruction method recently proposed by \citet{Tocci2011}. The reconstructions were generated from 3 virtual exposures of a simulated Kodak Kai-04050 sensor (used in our experimental HDR video system) captured 5 f-stops apart. The lowest exposed and the highest exposed sensors are perfectly aligned, while the middle sensor has a translational offset of $[0.4, 0.45]$ in pixel units. This is similar to the alignment errors reported by \citeauthor{Tocci2011} for their camera system. The method of \citeauthor{Tocci2011} copes rather well with translational misalignments, because each reconstructed pixel uses information from only one sensor. However, the reconstruction suffers from noise. This is expected as the method of \citeauthor{Tocci2011} only considers observations from the highest exposed non-saturated sensor at each point. In contrast LPA and CALPA are based on a maximum likelihood approach incorporating information from all non-saturated sensor observations. CALPA effectively denoises the image by using larger filtering supports in flat image areas. Compared to the other methods, CALPA also reduces unwanted color artifacts in areas of high spatial frequency.

\begin{figure*}[!t]
\begin{center}
\begin{subfigure}[b]{0.29\linewidth}
\includegraphics[width=\linewidth]{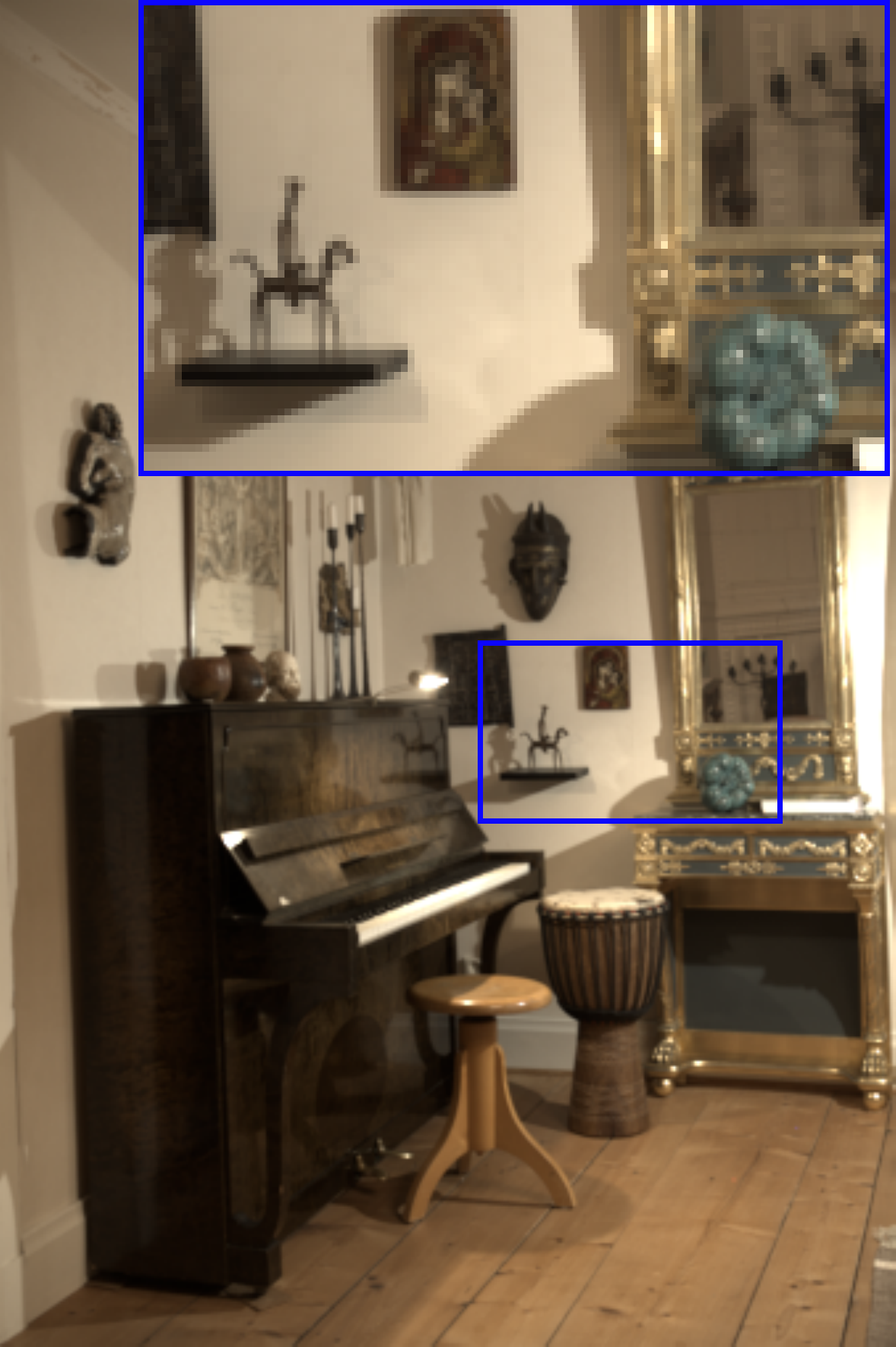}
\subcaption[b]{HDR Reference}
\end{subfigure}
\begin{subfigure}[b]{0.29\linewidth}
 \includegraphics[width=\linewidth]{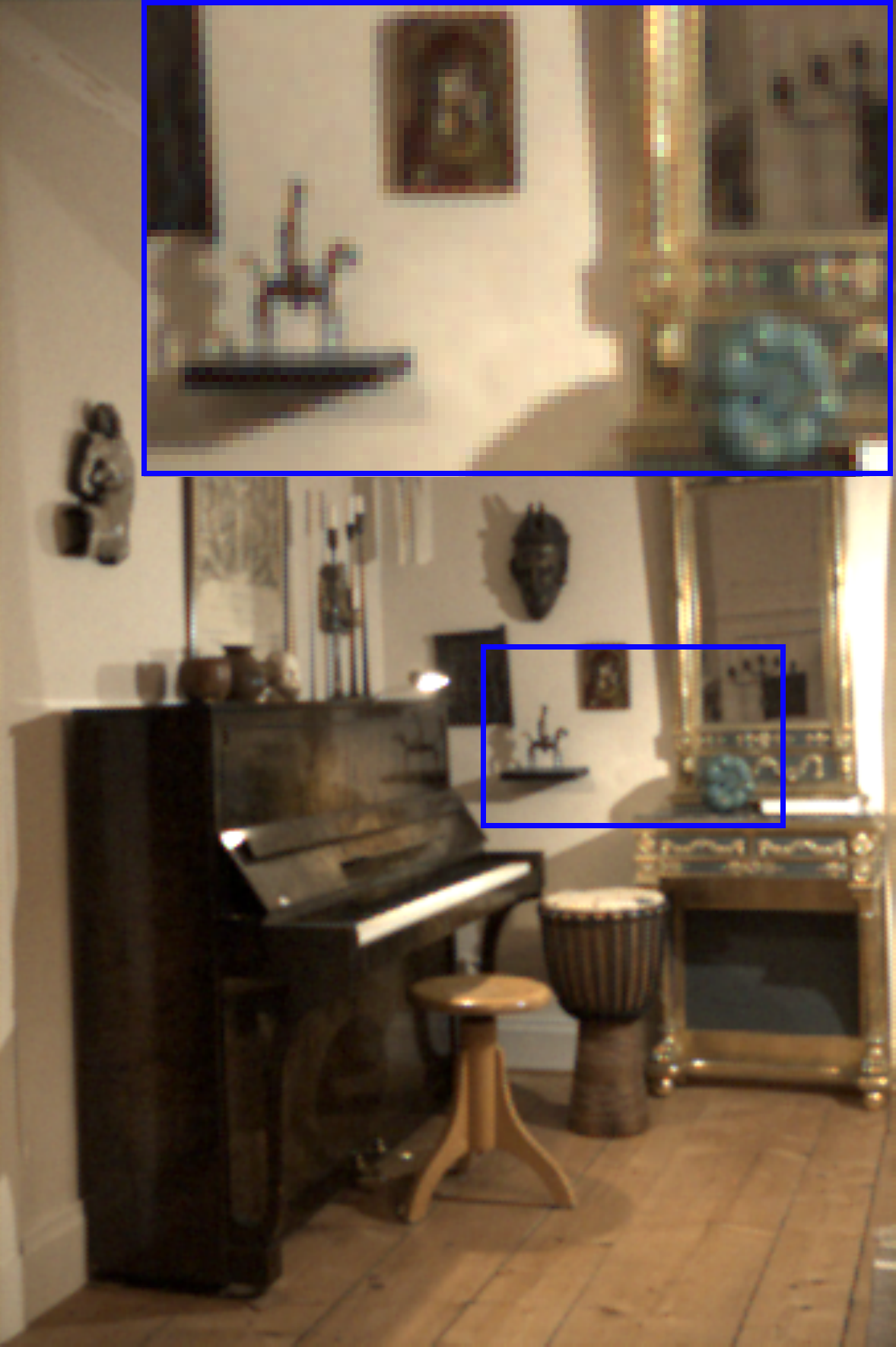}
\subcaption[b]{LPA, $M=1$}
\end{subfigure}
\begin{subfigure}[b]{0.29\linewidth}
 \includegraphics[width=\linewidth]{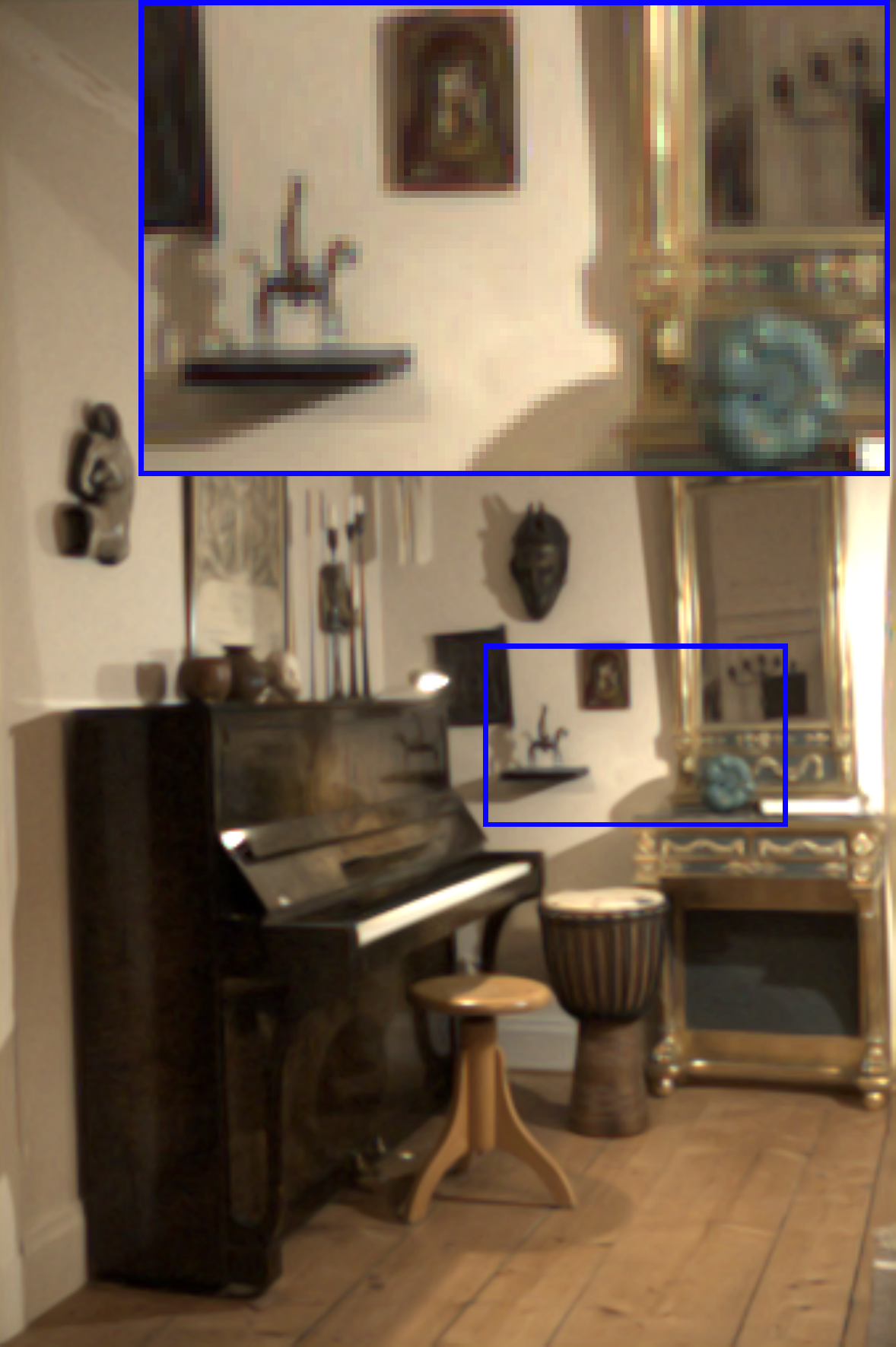}
\subcaption[b]{CALPA, $M=1$, $\alpha = 0.006$}
\end{subfigure}
\caption{Reconstructed results from 3 virtual sensors with $6$ degrees rotational misalignment between the sensors. a) Tonemapped HDR reference image (no bayer pattern). b) LPA, $M=1$ c)  CALPA, $M=1$, $\alpha = 0.006$. Results are best viewed close up in digital form.}
\label{fig:rotmissalignment}
\end{center}
\end{figure*}

A robust multi-sensor reconstruction algorithm requires handling of general sub-pixel misalignments between the sensors,
including rotational errors. Both LPA and CALPA gracefully handle any geometric misalignments between the sensors, as can be seen in Figure~\ref{fig:rotmissalignment}, where the reconstructions were based on 3 virtual Kodak Kai-04050 sensors simulated with 6 degree rotational errors between them. The Debayer-first method can also handle general misalignment errors including rotations. However, methods performing demosaicing before HDR assembly often suffer from color artifacts in regions of high spatial frequency, e.g. seen as color fringes in Figure~\ref{fig:PerfectAlignmentComparison}. Methods performing demosaicing after resampling (alignment), e.g. Tocci \etal ~\cite{Tocci2011}, cannot handle general misalignments, as the color filters no longer necessarily overlap after resampling. 

\begin{figure*}[!t]
\begin{center}
\begin{subfigure}[b]{0.2935\linewidth}
\includegraphics[width=\linewidth]{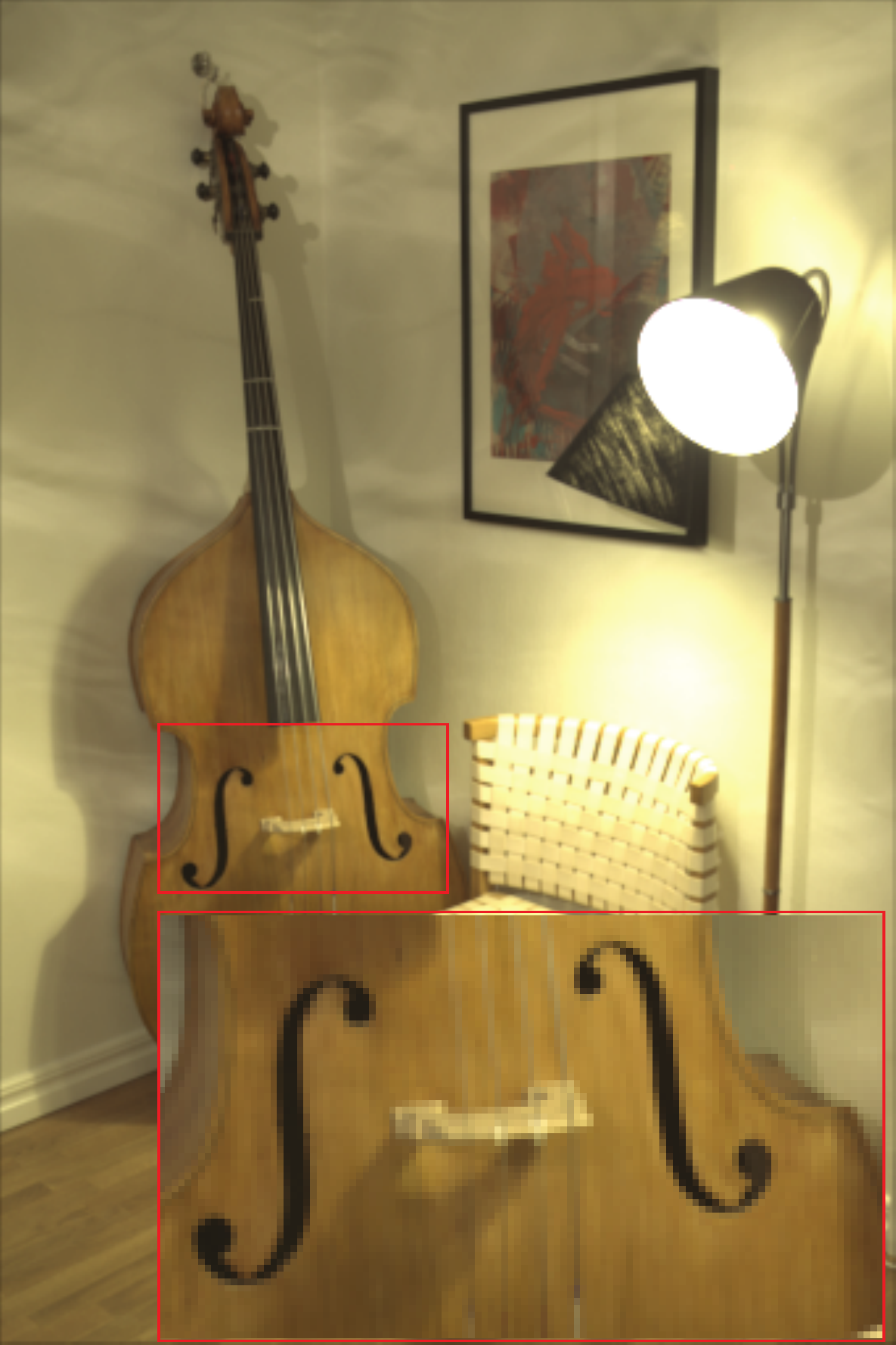}
\subcaption[b]{HDR Reference}
\label{subfig:ReferenceLPAORder}
\end{subfigure}
\begin{subfigure}[b]{0.25\linewidth}
 \includegraphics[width=\linewidth]{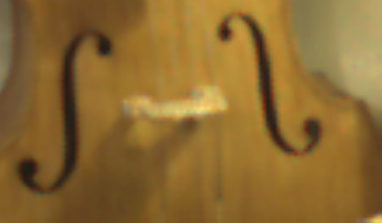}
 \includegraphics[width=\linewidth]{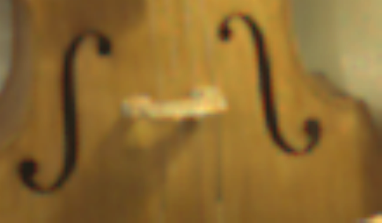}
 \includegraphics[width=\linewidth]{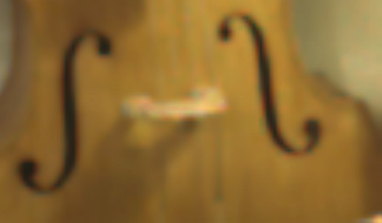}
\subcaption[b]{LPA, $M=1$, $h=0.3, 0.6, 0.9$}
\label{subfig:comparison3LPAM0}
\end{subfigure}
\begin{subfigure}[b]{0.25\linewidth}
 \includegraphics[width=\linewidth]{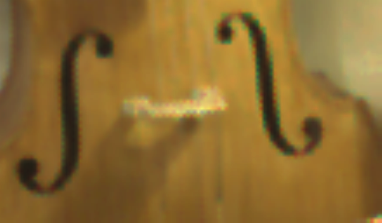}
 \includegraphics[width=\linewidth]{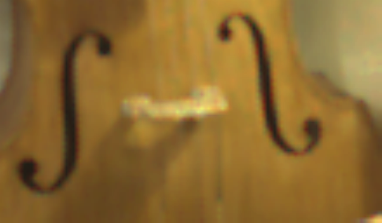}
 \includegraphics[width=\linewidth]{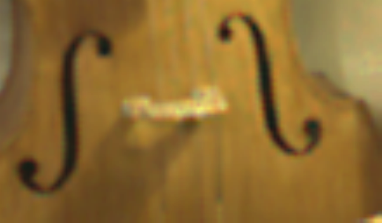}
\subcaption[b]{LPA, $M=0,1,2$, $h=0.4$}
\label{subfig:comparison3LPAM0}
\end{subfigure}
\caption{a) Tonemapped HDR reference image. b) LPA reconstructions, $M=1$, for different scale parameters, $h_{R,B} = 0.3, 0.6, 0.9$. Using a low scale parameter leads to sharp reconstructions. If severe image noise is present, a larger scale parameter may be beneficial. c)LPA reconstructions of different polynomial order $M=0,1,2$ with a fixed scale parameter, $h_G = 0.2$, $h_{R,B} = 0.4$. Using a constant polynomial ($M=0$), blocky artifacts and color fringes are observed close to sensor saturation and sharp edges.}
\label{fig:comparison3}
\end{center}
\end{figure*}

\subsection{LPA parameters}
Varying the polynomial order and the scale parameter for the LPA method imposes a trade-off between image sharpness, noise reduction and processing speed. Figure~\ref{fig:comparison3}b and \ref{fig:comparison3}c  show reconstructions from three aligned virtual Kodak KAI-04050 sensors 4 \textit{f}-stops apart and a resolution of $512 \times 341$ pixels. The close-ups show how the image fidelity varies for different choices of polynomial order, $M$, and the scale parameter, $h$, described in Section~\ref{subsec:parameters}. Using a constant fit, $M=0$, leads to unwanted block artifacts and color fringes, especially at edge features and in regions were one sensor is close to saturation. The difference between orders $M=1$ and $M=2$ is more subtle, with only small improvements in visual quality. A low value for the scale parameter, $h$, increases image sharpness, while a large $h$ reduces image noise.

\section{Experimental validation}
\label{sec:hdrcamera}
To evaluate the real world performance of our approach, we have implemented the LPA algorithm in CUDA to enable GPU  processing of the output from a custom built multi-sensor HDR camera system. 

\begin{figure}[!t]
\begin{center}
\includegraphics[width=\linewidth]{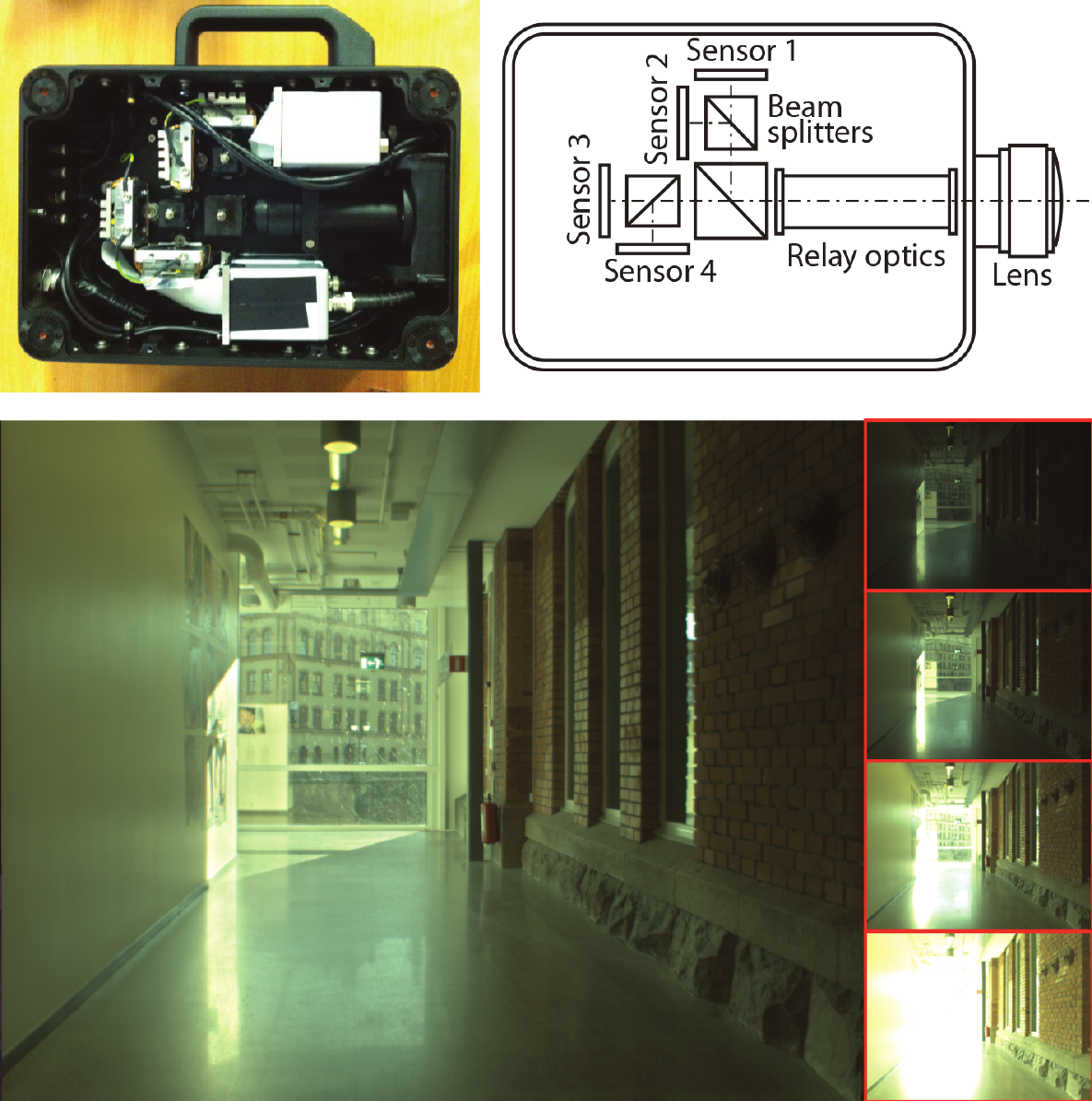} 
\caption{Top: An overview of our experimental HDR video camera. Bottom: a full resolution locally tonemapped frame from a video sequence captured with the HDR video camera. The image is reconstructed from four 4 Mpixel sensors at $26$ fps. Bottom-right: Four gamma mapped images $2$ f-stops apart from the same HDR-image.}
\label{fig:precomputeComparison}
\end{center}
\vspace{-20pt}
\end{figure}

The camera system used in our experiments is based on four high quality Kodak KAI-04050 CCD sensors with a resolution of $2336 \times 1752$ pixels and RGB Bayer pattern CFA sampling. The sensors receive different amounts of light from an optical system with a common lens, a four-way beam splitter and four different ND filters, see Figure~\ref{fig:precomputeComparison}, top. 
The relay lens was custom built for this setup, and extends the optical path length to make room for the beam splitters. All other optics are standard, off-the-shelf components. The sensors have 12 bits linear A/D conversion, and the exposure scalings cover a range of $1 : 2^{12}$, yielding a dynamic range equivalent to $12 + 12 = 24$ bits of linear resolution, commonly referred to as ``24 \textit{f}-stops'' of dynamic range. The dynamic range can be extended further by varying the exposure times between the sensors. The sensors are connected to a host computer through a CameraLink interface. The system allows for capture, processing and off-line storage of up to 32 frames per second at 4 Mpixels resolution, amounting to around 1 GiB/s per second of raw data.

The misalignments between the sensors were measured by imaging a checkerboard calibration target and computing the cross-sensor correlation using OpenCV 2.2. The misalignments could be characterized as translations of a few pixels and as 2D rotations in the order of fractions of a degree. The sensor transformation matrices could be estimated with an accuracy in the order of $0.1$ pixels. The system was radiometrically calibrated as described in Section~\ref{subsection:parametercalibration}.  The gain parameter  was estimated for each pixel separately, and then spatially averaged to find the sensor gain $g_{s} = 0.27 DV/e$.  Readout noise was estimated for each pixel, with a cross-sensor spatial mean of $\hat{\mu}_{s,i} = 72$ e, and a standard deviation of $\hat{\sigma}_{s,i} \approx 11.8$ e.

Our software implementation of the LPA algorithm required several seconds for processing of reasonably sized frames. To enable real time processing for high resolution video data, we implemented the LPA algorithm in CUDA. Two CUDA kernels are executed per frame. First, the raw sensor images are converted to a half-float radiant power estimate, $\hat{f}_{s,i}$, and a variance estimate, $\hat{\sigma}_{\hat{f}_{s,i}}$, is computed. Secondly, for each pixel in the output HDR image, all samples overlapping the observation window are read from global GPU memory. The corresponding weight is computed and used to compute the weighted least squares estimate described in Equation~\eqref{eq:wlsqe}. Since memory transfer is a bottle-neck, we use page-locked and write-combined memory to improve efficiency. The data is transferred to the GPU for kernel execution and display, and then back to the CPU for disk storage. To enable simultaneous data transfer and kernel execution
we utilize two CUDA streams.

If the sensors are aligned or related by purely translational transforms, and if the output resolution matches the input resolution, the spatial arrangement of samples in each observation window will be invariant across the image, and the window weights $w_i$ can be precomputed once. This results in a $2 \times$ to $3 \times$ speedup for a local constant fit $M=0$. 
 
Figure \ref{fig:precomputeComparison} shows an example frame captured using our experimental HDR-video system. Using our CUDA implementation with $M=0$, and an output resolution of $2336 \times 1752$ pixels, the  four input frames are processed at $26$ fps using pre-computed weights on an Nvidia GForce 680 GPU. As can be seen in the tonemapped image, which is in full resolution to enable zooming in during on-screen viewing, the small rotational misalignments present in the hardware setup are not visible even though the weights are pre-computed. This is partly due to the smoothing inherent in the LPA method. Reconstructing the same frame without pre-computed weights achieves an interactive speed of $7$ fps. The performance scales linearly with the number of pixels, so the corresponding figures for reconstructing to HD resolution frames of $1920 \times 1080$ are $51$ fps for pre-computed weights and $14$ fps without pre-computed weights.


\section{Conclusions and Future Work}
This paper presented a novel filtering framework for HDR reconstruction from multi-sensor images that performs all steps in the traditional HDR imaging pipeline in a single step. The reconstruction is based on spatially adaptive filtering and noise-aware local likelihood estimation using isotropic and anisotropic filter supports taking into account the correlation between color channels. The method uses a novel sensor noise model adapted to multi-sensor systems. We also presented an overview of a novel HDR video camera and showed how our algorithm achieves real-time performance. It should be noted that the framework is general in that the algorithms apply equally well to still image HDR reconstruction from e.g. exposure bracketing. The only requirement is that the sensor transformations can be accurately estimated.

As future work, we would like to extend the filter supports from 2D to 3D to include both the spatial and temporal domain in the reconstruction.






\bibliographystyle{elsarticle-num-names}
\bibliography{references}







\end{document}